\documentclass[10pt,twocolumn,letterpaper]{article}

\usepackage[pagenumbers]{iccv} 

\usepackage{graphicx}
\usepackage{amsmath}
\usepackage{amssymb}
\usepackage{booktabs}
\usepackage{float}
\usepackage{enumerate}
\usepackage[all]{nowidow}
\usepackage{uoftcolors}

\usepackage{mathtools} 
\usepackage{booktabs} 
\usepackage{tikz} 
\usepackage{amsfonts}
\usepackage{multirow}
\usepackage{multicol}
\usepackage{color,colortbl}
\usepackage{xspace}
\usepackage{wrapfig}
\usepackage{subcaption}
\usepackage{bbm}
\usepackage{bm}
\usepackage{url}
\usepackage{threeparttable} 

\def\eg{e.g.,\ }               
\def\ie{i.e.,\ }               

\definecolor{LavenderBlue}{rgb}{0.7020,    0.8039,    0.8902}
\definecolor{Lightapricot}{rgb}{0.9961,    0.8510,    0.6510}
\definecolor{thirdtablecolor}{rgb}{0.8706,    0.7961,    0.8941}

\newcommand{\heading}[1]{\noindent\textbf{#1}}

\long\def\ignorethis#1{}

\definecolor{demphcolor}{RGB}{100,100,100}

\newlength\pagetopmargin
\newlength\figcapmargin
\newlength\figmargin
\newlength\tablecapmargin
\newlength\tablemargin

\renewcommand{\arraystretch}{1.1}  

\usepackage{soul}

\newcommand{\R}{\mathbb{R}}
\newcommand{\mcL}{\mathcal{L}}
\newcommand{\img}{I}
\newcommand{\stream}{\mathcal{E}}
\newcommand{\polarityCnt}{E}
\newcommand{\intervalLength}{T}
\newcommand{\seqLength}{M}
\newcommand{\numEvents}{n}
\newcommand{\denormEvents}{N}
\newcommand{\bins}{B}
\newcommand{\contrast}{C}
\newcommand{\encoder}{\Phi_{\text{enc}}}
\newcommand{\decoder}{\Phi_{\text{dec}}}
\newcommand{\recurrent}{\Phi_{\text{rec}}}
\newcommand{\feats}{F}
\newcommand{\memory}{\bm{c}}
\newcommand{\realVideo}{L}
\newcommand{\intVideo}{\hat{L}}
\newcommand{\reconVideo}{\tilde{L}}
\newcommand{\stage}{\bm{s}}

\newcommand{\algoNameFull}{TESPEC\xspace}

\definecolor{iccvblue}{rgb}{0.21,0.49,0.74}
\usepackage[pagebackref,breaklinks,colorlinks,allcolors=iccvblue]{hyperref}

\def\paperID{6629} 
\def\confName{ICCV}
\def\confYear{2025}

\title{\algoNameFull: Temporally-Enhanced Self-Supervised Pretraining for Event Cameras}

\author{
Mohammad Mohammadi \textsuperscript{1, 2} \quad
Ziyi Wu \textsuperscript{1,2} \quad
Igor Gilitschenski \textsuperscript{1,2} \\
\textsuperscript{1}University of Toronto \quad
\textsuperscript{2}Vector Institute \\
{\tt\small \{mohammadi, ziyiwu, gilitschenski\}@cs.toronto.edu}
}

\begin{document}
\maketitle

\begin{abstract}
Long-term temporal information is crucial for event-based perception tasks, as raw events only encode pixel brightness changes.
Recent works show that when trained from scratch, recurrent models achieve better results than feedforward models in these tasks.
However, when leveraging self-supervised pre-trained weights, feedforward models can outperform their recurrent counterparts.
Current self-supervised learning (SSL) methods for event-based pre-training largely mimic RGB image-based approaches.
They pre-train feedforward models on raw events within a short time interval, ignoring the temporal information of events.
In this work, we introduce \algoNameFull, a self-supervised pre-training framework tailored for learning spatio-temporal information. \algoNameFull is well-suited for recurrent models, as it is the first framework to leverage long event sequences during pre-training.
\algoNameFull employs the masked image modeling paradigm with a new reconstruction target.
We design a novel method to accumulate events into pseudo grayscale videos containing high-level semantic information about the underlying scene, which is robust to sensor noise and reduces motion blur.
Reconstructing this target thus requires the model to reason about long-term history of events.
Extensive experiments demonstrate our state-of-the-art results in downstream tasks, including object detection, semantic segmentation, and monocular depth estimation.
Project webpage: \url{https://mhdmohammadi.github.io/TESPEC_webpage}.
\end{abstract}
\vspace{-6mm}
\section{Introduction}
\label{sec:intro}
\vspace{-2mm}
Event cameras are bio-inspired sensors that asynchronously record pixel intensity changes~\cite{EventVisionSurvey}. They offer distinct advantages, including low energy consumption, a high dynamic range, and high temporal resolution.
There has been growing interest in applying event cameras to various computer vision tasks~\cite{MVSEC, 1Mpx, Gen1, DDD17}.
However, this novel data modality also poses unique challenges, e.g., the need for specialized models.
Since individual events only encode short-term information, methods for complex tasks such as object detection usually aggregate events over a certain time interval ~\cite{EVCNNDet1, EVCNNDet2, EV-SegNet}.
Still, these \emph{feedforward} models discard long-horizon history, making it hard to capture objects under small motion that trigger very few events.
Recent models thus integrate recurrent modules~\cite{ConvLSTM} to utilize information beyond a relatively short time period~\cite{RVT, ESS, SSMforEvents, PCDepth, recurrenteventdepth1}, achieving superior performance.

Despite recent progress, performance on many event vision tasks is limited by a lack of large labeled datasets~\cite{MVSEC, DSEC, DDD17}. 
In conventional RGB vision, self-supervised learning (SSL) has proven effective for improving performance in data-scarce scenarios~\cite{SimCLR, MAE, MoCo}.
Therefore, several studies have introduced SSL to event-based vision~\cite{ECDP, MEM, ECDDP}, aiming to pre-train robust feature extractors on large-scale, unlabeled event data.
However, these approaches largely mimic conventional frame-based SSL.
They aggregate events over short time intervals into 2D image-like representations, and then perform contrastive learning~\cite{ECDP} or reconstruct masked locations~\cite{ECDDP, huang2024data, MEM}.
A core limitation here is that real-world event data is notably sparser than RGB images. 
When converting events to frames, many pixels remain empty or contain merely noise, which fails to provide a meaningful learning signal~\cite{ECDP}.
Overall, pre-training models to extract long-term information from events is an underexplored research problem.

We address this gap with \algoNameFull, a self-supervised pre-training framework designed to learn long-term information from event sequences.  
Our approach follows the masked image modeling (MIM) paradigm~\cite{MAE, SimMIM}, where the model receives partially masked event streams and is trained to reconstruct an unmasked target.
This requires the model to reason about current object locations based on their past motions.
Unlike RGB images, event data has a distinctive temporal dimension.
While raw events capture only low-level pixel brightness changes over short intervals, they can encode high-level semantic information when accumulated over longer periods.
Prior studies have shown that high-quality grayscale videos can be reconstructed from raw event streams~\cite{E2VID, E2VID-PAMI}.
Based on this insight, we propose using accumulated events to create pseudo-grayscale videos -- resembling grayscale frames -- as reconstruction targets.
Compared to the short-term event frames used in prior work, these pseudo videos provide richer long-term information and dense learning signals, supporting the training of recurrent architectures. 
In addition, approximating videos using events eliminates the need for paired video data in pre-training, making \algoNameFull a pure event-based self-supervised pre-training approach applicable to any event camera datasets.

A key challenge is how to obtain meaningful grayscale videos purely from event data.
An intuitive solution is to use pre-trained event-to-frame reconstruction models~\cite{E2VID, E2VID-PAMI}. This is infeasible in our SSL setting as such models require paired event-video training data.
An alternative approach would be using methods that estimate intensity signals by integrating events over time~\cite{brandli2014real, Intensity}.
However, we found their estimated videos suffer from severe motion blur when applied to outdoor sequences with high event rates.
This is because these methods process each pixel separately and fail to adapt to the global motion of the scene.
Object motion blur is particularly harmful to our pre-training approach, as it encourages the model to ``remember" previous object positions, which contrasts the objectives of downstream perception tasks.
To address this, we generalize the intensity estimation formulation from~\cite{Intensity} to process events in global batches, enhancing robustness against local noise.


In summary, this work makes two main contributions:
\textbf{(i)} We highlight an unexplored instance of event-based SSL that explicitly encourages the model to learn long-term temporal information, 
\textbf {(ii)} We propose an improved intensity video representation as the pre-training objective, leading to enhanced temporal information learning.
As a consequence, our approach achieves state-of-the-art performance on multiple downstream event-based perception tasks.

\vspace{-2.5mm}
\section{Related Work}
\label{sec:related-work}
\vspace{-1.5mm}
\heading{SSL for RGB images} can be mostly categorized into three classes.
Earlier works adopt the contrastive learning framework~\cite{SimCLR, SimCLRV2, MoCo, MoCoV2, MoCoV3, SwAV, SimSiam, InstanceDiscrimination, ExemplarLearning, CPC, CPCV2, CMC}.
The model is trained to perform instance discrimination by pushing two views of the same data closer, while separating views from different samples.
Another line of work performs self-distillation~\cite{DINO, DINOV2, BYOL, iBOT}. 
These methods only compare the model features on the same data, eliminating the need for negative samples.
Recent years have witnessed the renaissance of reconstruction-based pre-training~\cite{AE, StackedAE, ColoringSSL, InpaintingSSL, AAE}. 
Inspired by the success of large language models~\cite{BERT, GPT-1, GPT-2}, recent works propose masked image modeling (MIM) as the pre-training objective, where a model takes in a partially masked image, and is trained to reconstruct the unmasked one.
Different reconstruction targets have been explored in the literature, including raw pixel values~\cite{MAE, SimMIM, i-GPT}, discrete indices from a pre-trained tokenizer~\cite{BEiT, BEiTV2, BEiTV3, PeCo}, and intermediate features~\cite{MaskedFeat, CAE, EVA}.
The representative work MAE~\cite{MAE} shows that ViTs~\cite{ViT} can reconstruct input images even with 75\% of pixels masked out.
In this work, we adopt the SSL framework similar to MAE and design a new reconstruction target tailored for event camera data.

\heading{MAE for RGB videos.}
Several works have extended the MAE framework to video data~\cite{VideoMAENJU, VideoMAEKaiming, VideoMAEV2}.
Compared to images, a distinct property of videos is temporal dynamics.
Thus, some research incorporates motion into video MAE, such as guiding pixel masking with object motion~\cite{MotionGuidedMAE1, MotionGuidedMAE2}, predicting temporal differences between frames~\cite{MotionMAEFrameDiff1, MotionMAEFrameDiff2}, or predicting object trajectories~\cite{MotionMAETraj, MotionMAE4ActionCls}.
Interestingly, prior works observed degraded performance when only predicting frame differences~\cite{MotionMAEFrameDiff1, MotionMAEFrameDiff2, MotionMAE4ActionCls}.
Based on this insight, we propose reconstructing the accumulated event video instead of raw events as the pre-training objective.
This forces the model to extract long-term temporal information from data.

\heading{SSL for event camera data.}
Due to a lack of large labeled datasets, many works have studied label-efficient learning on event data~\cite{ESS, EventDA, EventBind, EventCLIP, LEOD, Ev-LaFOR, EventGraftNet, EventDance}.
Some design task-specific constraints for unsupervised learning~\cite{zhu2019unsupervised, EvSSLReconRGB, EvSSLDescriptor, EvSSLOptFlowSNN, EvSSLVideoDerain, zhang2022data}.
Instead, this paper aims to pre-train a versatile backbone.
ECDP~\cite{ECDP} trains the model to align features extracted from paired RGB images and events.
MEM~\cite{MEM} and Huang et al.~\cite{huang2024data} follow MAE to reconstruct input events.
The state-of-the-art ECDDP~\cite{ECDDP} designs sophisticated losses combining self-distillation and MIM.
Notably, all these methods are designed for \emph{feedforward} models, while \emph{recurrent} models have shown better performances in the event vision literature~\cite{RVT, SSMforEvents, ESS, E-FlowFormer}.
Our work is the first event SSL framework for pre-training recurrent models, which achieves state-of-the-art results on downstream tasks with a temporally-aware learning objective.

\heading{Recurrent architectures in event-based vision.}
To make a prediction, feedforward models only process events within a recent time range~\cite{EV-SegNet, EVCNNDet1, EVCNNDet2, EVCNNDet3}.
However, raw event data encode pixel value changes and thus contain limited information.
For tasks requiring long-term memory, recent works introduce recurrency to their backbones, including object detection~\cite{ASTMNet, RVT, 1Mpx}, semantic segmentation~\cite{ESS}, optical flow~\cite{E-RAFT, E-FlowFormer}, and depth estimation~\cite{PCDepth, recurrenteventdepth1, recurrenteventdepth2}.
They usually add recurrent modules~\cite{LSTM, ConvLSTM} in the backbone, enabling reusing information from the last event segment.
We follow this trend and propose \algoNameFull to learn a strong recurrent backbone to benefit these downstream tasks.

\vspace{-2mm}
\section{Method}
\label{sec:method}
\vspace{-1.5mm}
\begin{figure*}[t]
    \vspace{\pagetopmargin}
    \vspace{-4mm}
    \centering
    \begin{subfigure}{1.0\textwidth}
        \includegraphics[width=1.0\textwidth]{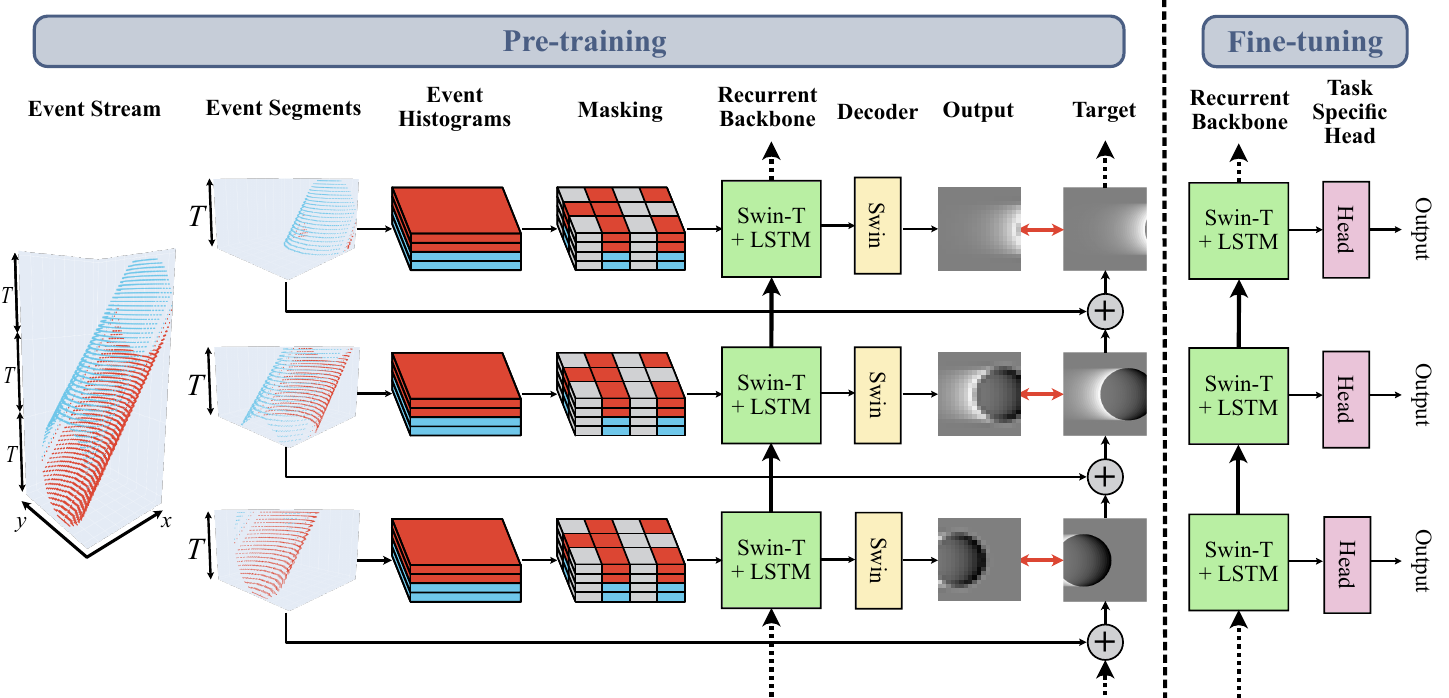}
        \vspace{-4mm}
    \end{subfigure}
    \vspace{\figcapmargin}
    \vspace{-5mm}
    \caption{
        \textbf{Overview of the \algoNameFull Pipeline.}
        \emph{Pre-training} (left):
        Given a raw event stream (red and blue represent negative and positive events, respectively), we first split it into non-overlapping segments, and convert each segment into an event histogram.
        They are used to update the estimated intensity video of the scene.
        Then, we apply temporal tube masking to event histograms that masks out the same spatial locations across time.
        The masked input is fed in our recurrent backbone, which extracts features by fusing history information from previous events and current inputs.
        With the extracted features, a lightweight feedforward decoder reconstructs the intensity video at masked patches.
        \emph{Fine-tuning} (right):
        After pre-training, we discard the decoder and attach a new task-specific head to the recurrent backbone.
        Then, the whole model is fine-tuned on the downstream dataset.
    }
    \label{fig:pipeline}
    \vspace{\figmargin}
    \vspace{-3mm}
\end{figure*}

\algoNameFull adopts the MAE~\cite{MAE} framework to pre-train a backbone for event camera data (\cref{sec:background-mae}).
We construct intensity videos from raw event streams that contain rich temporal information (\cref{sec:event-repr} \& \cref{sec:recon-target}).
The pre-training objective is a masked event reconstruction loss (\cref{sec:loss-fn}).
The overall pipeline of \algoNameFull is illustrated in \cref{fig:pipeline}.

\subsection{Background}
\label{sec:background-mae}

\heading{MAE and VideoMAE.} MAE~\cite{MAE} performs masked image reconstruction with an asymmetric encoder-decoder architecture.
Given an input image $\img$, it is first divided into non-overlapping patches.
Then, a high proportion of patches is randomly masked out, and the remaining ones are fed into a ViT~\cite{ViT} encoder to extract visual features.
Finally, it appends equal amounts of learnable tokens to the encoded features and runs a ViT decoder to reconstruct masked patches.
The training loss is a simple MSE over the masked patches:
\vspace{-1.5mm}
\begin{equation}
\label{eq:mae-loss}
    \mcL_{\text{MAE}} = \frac{1}{|\Omega|} \sum_{(x, y) \in \Omega} \left\| I(x, y) - \hat{I}(x, y) \right\|^2,
\vspace{-1.5mm}
\end{equation}
where $\Omega$ is the subset of masked patches, and $\hat{I}$ is the reconstructed image.
One key design choice in MAE is making the decoder significantly smaller than the encoder, leaving the encoder fully responsible for extracting visual features.

Later works extend MAE to video data~\cite{VideoMAENJU, VideoMAEKaiming} with the same reconstruction-based learning objective.
Compared to images, videos have a distinct property of temporal correlations between frames.
Therefore, Tong et al.~\cite{VideoMAENJU} propose tube masking, which masks out patches at the same spatial location throughout the entire video. This prevents information leaks and makes the pre-training task more challenging.


\heading{Event-Camera Data.} Event cameras capture per-pixel log-intensity changes, and produce a stream of events $\stream = \{e_i = (x_i, y_i, t_i, p_i)\}$.
An event $e_i$ is triggered at time $t_i$ when the log-intensity at pixel $(x_i, y_i)$ changes (since the last event at this location) beyond a pre-defined threshold $\contrast$, i.e. when
\begin{equation}
\label{eq:event-camera-mechanism}
    \realVideo(x_i, y_i, t_i) - \realVideo(x_i, y_i, t_i - \Delta t) = p_i \contrast,
\end{equation}
where $p_i \in \{-1, 1\}$ is the event polarity, and $\Delta t$ is the time elapse since the last event at $(x_i, y_i)$.

\vspace{-0.75mm}
\subsection{Event Processing for Temporal MAE}
\label{sec:event-repr}
\vspace{-0.5mm}
An MAE is not directly applicable to event data due to the event's sparse and asynchronous nature.
To bridge the modality gap, we convert raw events to 2D frames.
We adopt the event histogram representation~\cite{RVT} due to its simplicity and good performance in prior works~\cite{EVCNNDet1, EVCNNDet2, EVCNNDet3}.
We first split the event stream $\stream$ into $\seqLength$ non-overlapping segments $\{\stream_i\}_{i=1}^\seqLength$, each covering events in a fixed time interval $\intervalLength$.
Then, we create a 4D tensor $\stage_i \in \R^{2 \times \bins \times H \times W}$ (dubbed stage) for each $\stream_i$ as follows:
\begin{align}
\label{eq:event-repr}
    \stage_i(p, \tau, x, y) & = \sum_{e_j \in \stream_i}\delta(p - p_j)\delta(x-x_j, y-y_j)\delta(\tau - \tau_j), \notag \\
    \tau_j & = \left\lfloor\left(\frac{t_j}{\intervalLength} - i + 1\right) \times \bins\right\rfloor,
\end{align}
where $\delta$ is the Dirac delta function.
Intuitively, each stage $\stage_i$ further divides the event segment $\stream_i$ into $\bins$ temporal bins, and counts the number of positive and negative events separately in each pixel.
Finally, we flatten the first two dimensions of $\stage_i$, resulting in a multi-channel image-like representation with shape $(2\bins, H, W)$.

\begin{figure}[t]
    \vspace{-6mm}
    \centering
    \begin{subfigure}{0.32\linewidth}
        \includegraphics[width=1.0\linewidth]{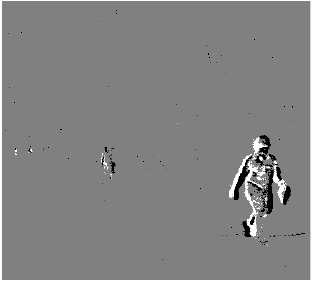}
        \vspace{-4mm}
        \caption{Input events within a short time interval.}
    \end{subfigure}
    \hspace{-0.5mm}
    \begin{subfigure}{0.32\linewidth}
        \includegraphics[width=1.0\linewidth]{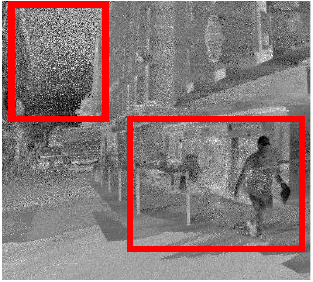}
        \vspace{-4mm}
        \caption{Estimated intensity, prior method (\cref{eq:prior-intensity-est}).}
    \end{subfigure}
    \hspace{-0.5mm}
    \begin{subfigure}{0.32\linewidth}
        \includegraphics[width=1.0\linewidth]{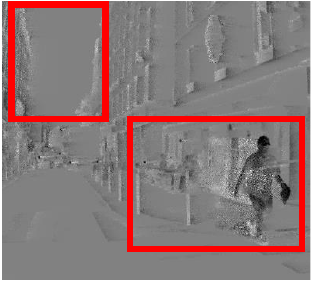}
        \vspace{-4mm}
        \caption{Estimated intensity, our method (\cref{eq:our-intensity-est}).}
    \end{subfigure}
    \vspace{\figcapmargin}
    \caption{
        \textbf{Visualization of the estimated intensity on a real-world example.}
        (a) Input events in a short interval are typically sparse.
        (b) The estimated image with the prior method contains motion blur behind the pedestrian, and severe sensor noise on the static background.
        (c) By considering events globally, our method produces images with sharp object motion and clean backgrounds.
    }
    \label{fig:grayscale-img-real-world}
    \vspace{-5mm}
\end{figure}

\heading{Vanilla event MAE.}
Given the 2D event frame $\stage_i$, prior works~\cite{huang2024data, MEM} simply mask it and run a feedforward model to do reconstruction.
However, $\stage_i$ only contains events in a short range and is usually sparse as shown in \cref{fig:grayscale-img-real-world} (a).
Reconstructing it is inefficient as many pixels are empty or contain only noisy events.
In addition, events are mostly triggered at object boundaries.
Forcing the model to reconstruct edges instead of the whole object may harm downstream dense prediction tasks such as depth estimation.

\heading{Temporally-enhanced event MAE.}
We aim to construct a representation from raw events that contain rich learning signals.
Our key insight is that, while events in each time segment $\stream_i$ present only low-level brightness changes, accumulated events over multiple segments can encode high-level semantic information.
Indeed, while many objects could be static within a short segment (\eg cars stop at the traffic light), they should move and trigger events when observed for longer.
We thus propose integrating the entire event sequence $\stream$ to obtain a video $\intVideo \in \R^{H \times W \times \seqLength\intervalLength}$ as:
\begin{equation}
\label{eq:accum-events-high-level}
\hspace{-2mm}
    \intVideo(x, y, i \intervalLength) = \mathrm{Int}(\intVideo(x, y, (i-1) \intervalLength), \stream_i),\ \  \intVideo(\cdot, \cdot, 0) = 0,
\end{equation}
where $\mathrm{Int}(\cdot, \cdot)$ is the integration function to be discussed in \cref{eq:our-intensity-est}.
As we show in \cref{sec:recon-target}, $\intVideo$ resembles the grayscale video of the scene and thus contains semantic information.

The multi-stage integration objective also aligns with the recurrent architectures in event vision tasks.
These models typically take in event stages $\{\stage_i\}_{i=1}^\seqLength$ sequentially, and use recurrent modules~\cite{ConvLSTM} to carry information over time:
\begin{equation}
\label{eq:recurrent-model}
    \feats_i = \encoder(\memory_{i-1}, \stage_i),\ \  \memory_i = \recurrent(\memory_{i-1}, \stage_i),\ \ \memory_0 = \mathbf{0}.
\end{equation}
where $\encoder$ is the backbone and $\feats_i$ is the extracted feature, $\recurrent$ is the recurrent module and $\memory_i$ is its memory state.
Comparing \cref{eq:recurrent-model} to \cref{eq:accum-events-high-level}, we find similar update rules between the output feature $\feats_i$ and the accumulated video $\intVideo_i$.
Therefore, by reconstructing $\intVideo_i$ from $\feats_i$, we can learn a backbone that extracts features with history information.

\begin{figure}[t]
    \vspace{-6mm}
    \centering
    \begin{subfigure}{0.32\linewidth}
        \includegraphics[width=1.0\linewidth]{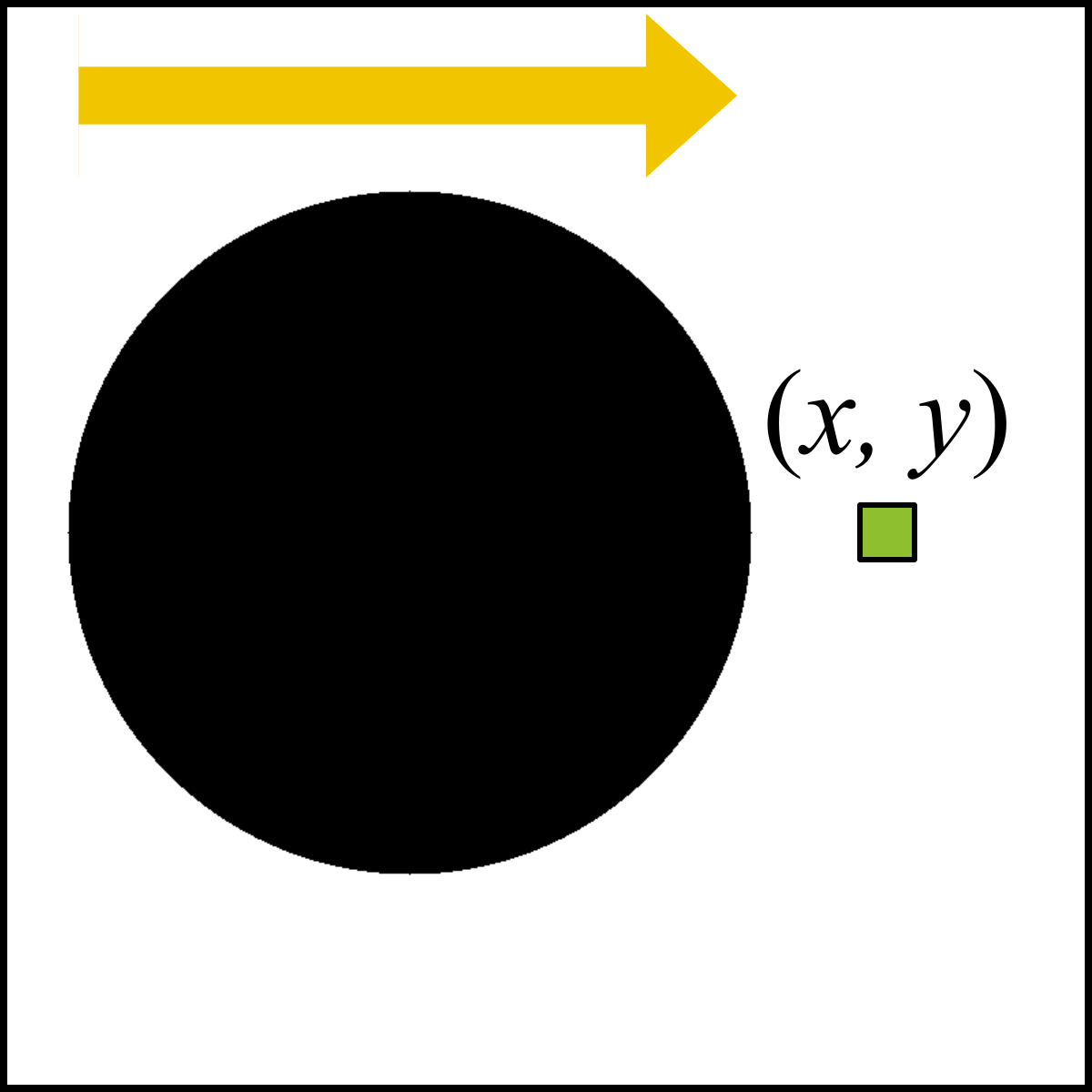}
        \vspace{-4mm}
        \caption{Toy example.}
    \end{subfigure}
    \hspace{-0.5mm}
    \begin{subfigure}{0.32\linewidth}
        \includegraphics[width=1.0\linewidth]{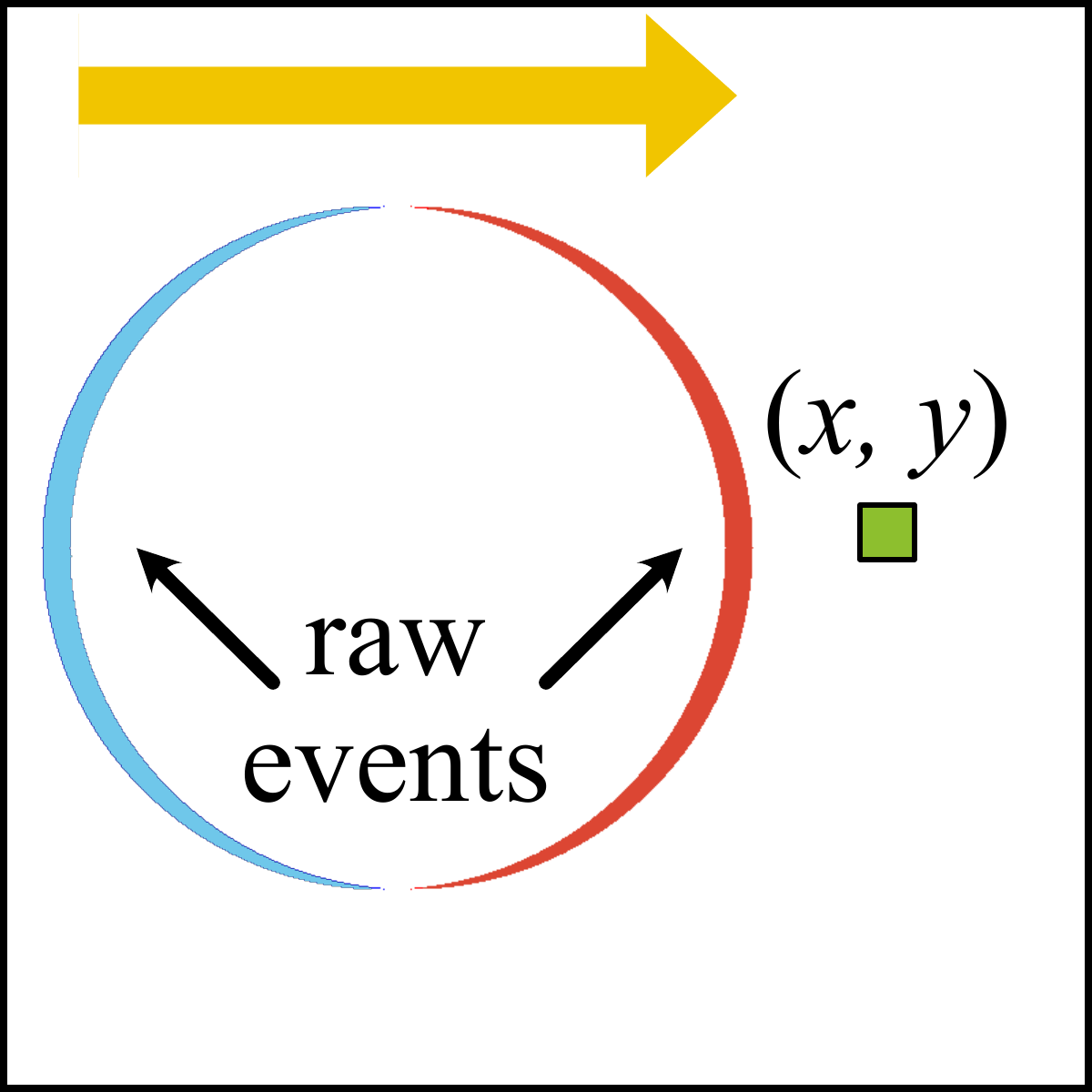}
        \vspace{-4mm}
        \caption{Triggered events.}
    \end{subfigure}
    \hspace{-0.5mm}
    \begin{subfigure}{0.32\linewidth}
        \includegraphics[width=1.0\linewidth]{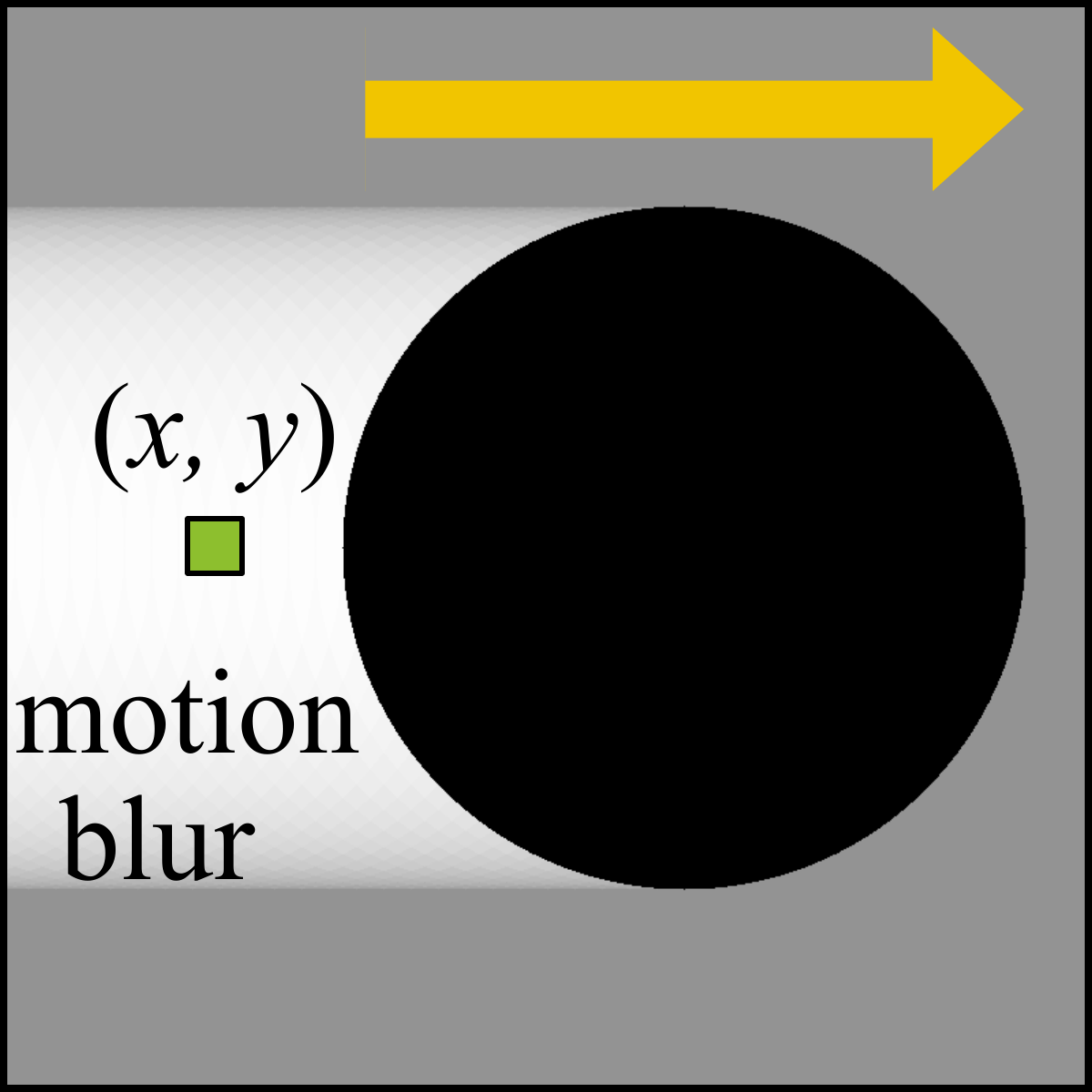}
        \vspace{-4mm}
        \caption{Estimated intensity.}
    \end{subfigure}
    \vspace{\figcapmargin}
    \caption{
        \textbf{Visualization of the estimated intensity image using \cref{eq:prior-intensity-est} on a toy example.}
        (a) A black circle moves on a white background.
        (b) Assuming no sensor noise, it will trigger one negative and one positive event when passing through a pixel $(x, y)$.
        (c) Due to the temporal decay term in \cref{eq:prior-intensity-est}, the events cannot cancel out, and thus will leave motion blur on the estimated image.
    }
    \label{fig:grayscale-img-toy-ex}
    \vspace{-5mm}
\end{figure}

\subsection{Improving the MAE Reconstruction Target}
\label{sec:recon-target}

According to \cref{eq:event-camera-mechanism}, given the initial grayscale image of a scene, one can reconstruct the real video $\realVideo$ by integrating events per pixel over time.
This is, however, impractical in our pure event-based SSL setting because it lacks image ground truth.
Additionally, real-world event cameras often exhibit considerable sensor noise.
Naive integration will quickly lead to visual quality degradation due to error accumulation~\cite{E2VID-PAMI}.
Some studies thus incorporate temporal filtering to achieve robust intensity estimation from raw events~\cite{Intensity, brandli2014real}.
We build upon the formula in 
\cite{Intensity}:
\begin{equation}
\label{eq:prior-intensity-est}
    \intVideo(x_i, y_i, t_i) = \exp{(-\alpha \Delta t)} \cdot \intVideo(x_i, y_i, t_i - \Delta t) + p_i \contrast.
\end{equation}
Intuitively, \cref{eq:prior-intensity-est} updates the value of each pixel $(x_i, y_i)$ individually.
When a new event $e_i$ comes at time $t_i$, it first performs an exponential decay on the previous state, and then increases or decreases the pixel value according to its polarity $p_i$.
The decay term suppresses noise accumulation and is crucial to obtain a stable estimation.

However, the estimated image from \cref{eq:prior-intensity-est} is prone to motion blur.
Consider a toy example shown in \cref{fig:grayscale-img-toy-ex} (a), where a black circle moves over a pixel $(x, y)$ on a white background.
When there is no sensor noise, the circle will only trigger one positive event and one negative event when it hits and leaves the pixel (\cref{fig:grayscale-img-toy-ex} (b)).
Without the decay term in \cref{eq:prior-intensity-est}, the positive event will first increase the pixel value by $C$, and then get canceled out by the negative event, resuming the original pixel value.
Now with temporal decay, the two events cannot cancel out and the final pixel value becomes $(e^{-\alpha \Delta t} - 1) \cdot \contrast$.
As shown in \cref{fig:grayscale-img-toy-ex} (c), this causes motion blur along the trajectory of the moving object and gets worse when the time elapse $\Delta t$ is larger.

In real-world event streams, this is a common scenario, \eg when a pedestrian walks across the street, the walking speed is small, and thus $\Delta t$ is large, leading to severe motion blur as shown in \cref{fig:grayscale-img-real-world} (b).
Having motion blur in the reconstruction target is problematic, as it encourages the model to remember all past locations of the object.
This is a waste of model capacity since downstream tasks often only focus on the current location of objects.
In addition, real-world event cameras have sensor noise such as hot pixels that keep producing events of the same polarity, leading to severe noise on the estimated grayscale image.

\heading{Our solution.}
The issue with \cref{eq:prior-intensity-est} is that each pixel is modeled separately.
The intensity of a pixel is never updated if it does not receive new events, making the motion blur persist.
To address it, we propose to update \emph{all pixels} whenever the camera receives new events.
Notably, instead of updating $\intVideo$ on every incoming event, we assume all events within a temporal bin in $\stream_i$ share the same timestamp, and update $\intVideo$ based on each bin.
This enables batch computation of our formula, which is fast on GPUs.
Additionally, we adjust the exponential decay term with the number of received events.
The new update rule is as follows:
\vspace{-0.5mm}
\begin{align}
\label{eq:our-intensity-est}
    \intVideo\left(x, y, t\right) & = \exp \left(-\alpha \Delta t \times \frac{\numEvents}{\denormEvents}\right) \cdot \intVideo\left(x, y, t - \Delta t\right) \notag \\
    & + \polarityCnt(x, y, t, \Delta t)\cdot \contrast, \ \ \text{where}\ \ \Delta t = \frac{\intervalLength}{\bins}.
\end{align}
\vspace{-0.5mm}
Here, the time elapse $\Delta t$ is always the interval of a temporal bin, $\frac{\intervalLength}{\bins}$, $\polarityCnt(x, y, t, \Delta t)$ is the signed accumulation of events at pixel $(x, y)$ within the temporal bin, and $\denormEvents$ is a fixed normalization factor.
Importantly, the temporal decay factor is now positively correlated to the number of events $\numEvents$ triggered during the entire $\Delta t$.
Consider the example in \cref{fig:grayscale-img-toy-ex} again, although the signed event count $\polarityCnt(x, y, t, \Delta t)$ remains zero after the circle leaves, the pixel value at $(x, y)$ keeps getting decayed, thus eliminating the motion blur.
This is verified by the real-world result in \cref{fig:grayscale-img-real-world} (c), where objects and scene elements are much sharper.
In addition, the noisy pixels on the background are also suppressed.

An interesting question is why involving the number of events $\numEvents$ in the exponential decay term---even if we remove $\numEvents$ from \cref{eq:our-intensity-est}, the motion blur will still disappear over time.
However, consider an extreme case where all objects are static and no events are triggered, $\intVideo$ remains unchanged in \cref{eq:our-intensity-est} since $\numEvents = 0$, which is desired.
In contrast, $\intVideo$ will be decayed without considering $\numEvents$.
Intuitively, the absence of new events suggests that the intensity should remain relatively unchanged, whereas a high number of events indicates sufficient information, allowing old events to be forgotten without compromising the estimation.
Therefore, we use $\numEvents$ to control the decay speed of history.
\subsection{Event-based Temporal MAE Pre-training}
\label{sec:loss-fn}
\vspace{-0.25mm}
Given an event stream $\stream$, we first convert it to a sequence of stages $\{\stream_i\}_{i=1}^\seqLength$ using \cref{eq:event-repr} and divide each stage into patches.
Then, we apply tube masking which masks out the same spatial patches across all stages in one training step, following VideoMAE~\cite{VideoMAENJU}.
The masked data is fed to the recurrent backbone sequentially to obtain features $\{\feats_i\}_{i=1}^\seqLength$ following \cref{eq:recurrent-model}.
On the other hand, we construct a grayscale video $\intVideo$ from $\stream$ using \cref{eq:our-intensity-est}.
We leverage a lightweight feedforward decoder head $\decoder$ that takes in event features extracted by the backbone, $\{\feats_i\}_{i=1}^\seqLength$, and predicts a video $\{\reconVideo_i \in \R^{H \times W}\}_{i=1}^\seqLength$.
The MSE between the estimated and the predicted video serves as our main loss,
\vspace{-0.5mm}
\begin{align}
\label{eq:recon-loss}
    \mcL_{\text{pretrain}} = \frac{1}{\seqLength \cdot |\Omega|} \sum_{i=1}^\seqLength & \sum_{(x, y) \in \Omega} \left\| \reconVideo_i(x, y) - \intVideo(x, y, i \intervalLength) \right\|^2, \notag \\
    \reconVideo_i & = \decoder(\feats_i).
\end{align}
A lightweight feedforward decoder forces the encoder to learn the spatio-temporal interactions in event sequences.

After pre-training, we discard the decoder, and attach a task-specific head to the recurrent backbone, which are jointly fine-tuned on the downstream dataset for each task.

%


\begin{table}[t]
    \vspace{\pagetopmargin}
    \centering
    \setlength{\tabcolsep}{2.8pt}
    \setlength{\arrayrulewidth}{0.5pt} 
    \rowcolors{4}{gray!20}{white}
    \footnotesize
  \begin{tabular}{l c c c c c}
    \toprule
    \multirow{2}{*}{Method} & \multirow{2}{*}{Backbone} & \multirow{2}{*}{Recurrent} & \multirow{2}{*}{Pre-training} & \multicolumn{2}{c}{mAP $\uparrow$} \\
    \cmidrule(lr){5-6}
     &  &  &  & Gen1 & 1Mpx\\
    \midrule
    \multicolumn{6}{l}{\textit{The best performance in the literature.}} \\
    FCCO~\cite{FCCO} & SwinV2 & No & MS-COCO & 50.4 & 40.6 \\
    RVT~\cite{RVT} & MaxViT & Yes & - & 47.2 & 47.4 \\ 
    GET-T~\cite{Get-T} & Transformer & Yes & - & 47.9 & 48.4 \\ 
    \hline 
    No Pre-training & Swin-T/7 & No & - & 36.2 & 34.2\\  
    RGB Supervised & Swin-T/7 & No & ImageNet-1k & 39.1 & 38.5 \\
    MoBY~\cite{MoBY} & Swin-T/7 & No & ImageNet-1k & 41.0 & 36.9\\  
    \hline
    No Pre-training & Swin-T/7 & Yes & - & 48.5 & 45.4 \\  
    RGB Supervised & Swin-T/7 & Yes & ImageNet-1k & 50.5 & 48.9 \\
    MoBY~\cite{MoBY} & Swin-T/7 & Yes & ImageNet-1k & 50.2 & 48.2\\
    \hline
    ECDDP~\cite{ECDDP} & Swin-T/7 & No & E-TartanAir & 40.8  & 38.5 \\
    ECDDP~\cite{ECDDP} & Swin-T/7 & Yes & E-TartanAir & 49.6 & 50.0 \\  
    \hline
    \algoNameFull (\textbf{Ours}) & Swin-T/7 & Yes & 1Mpx & \textbf{51.6} & \textbf{50.6} \\  
    \bottomrule
  \end{tabular}
  \vspace{\tablecapmargin}
  \caption{
  Object detection results on Gen1~\cite{Gen1} and 1Mpx~\cite{1Mpx} datasets.
  Best results are highlighted in \text{bold}.
  We report mean average precision (mAP) for evaluation.
  \algoNameFull achieves state-of-the-art results on both Gen1 and 1Mpx, surpassing all existing SSL methods in the RGB and event data domains.
  }
  \vspace{\tablemargin}
  \vspace{-4mm}
  \label{tab:obj-det}
\end{table}

\begin{table*}[t]
  \vspace{\pagetopmargin}
  \vspace{-4mm}
  \centering
  \setlength{\tabcolsep}{10.8pt}
  \setlength{\arrayrulewidth}{0.5pt} 
  \rowcolors{6}{white}{gray!20}
  \begin{tabular}{lccccccc}
    \toprule
    \multirow{2}{*}{Method} & \multirow{2}{*}{Backbone} & \multirow{2}{*}{Recurrent} & \multirow{2}{*}{Pre-training} & \multicolumn{2}{c}{DSEC}  & \multicolumn{2}{c}{DDD17} \\ 
    \cmidrule(lr){5-6}  \cmidrule(lr){7-8}
     &  &  &  & mIoU $\uparrow$ & mAcc $\uparrow$ & mIoU $\uparrow$ & mAcc $\uparrow$ \\
    \midrule
    \multicolumn{3}{l}{\textit{The best performance in the literature.}}  \\
    ESS~\cite{ESS} & - & Yes & - &  53.295 & 62.942 & 61.370 & 70.874 \\ 
    \hline
    No Pre-training & Swin-T/7 & No & - & 53.665 & 61.194 & 53.996 & 64.591 \\  
    RGB Supervised & Swin-T/7 & No & ImageNet-1k & 59.680 & 66.823 & 59.324 & 70.625 \\  
    MoBY~\cite{MoBY} & Swin-T/7 & No & ImageNet-1k & 58.553 & 66.070  & 57.667 & 67.639 \\ 
    \hline
    No Pre-training & Swin-T/7 & Yes & -  & 54.275 & 61.474 & 55.811 & 66.766  \\ 
    RGB Supervised & Swin-T/7 & Yes & ImageNet-1k  & 60.867 & 68.258 &  60.800 & 70.670 \\  
    MoBY~\cite{MoBY} & Swin-T/7 & Yes & ImageNet-1k & 59.709 & 67.220 & 59.970 & 69.284 \\  
    \hline 
    ECDP~\cite{ECDP} & ResNet50 & No & N-ImageNet & 59.155 & 67.534 & 59.145 & 70.176 \\ 
    ECDDP$^\dagger$~\cite{ECDDP} & ResNet50 & No & E-TartanAir & 60.641 & 69.502 & 62.912 & 74.015 \\ 
    ECDDP$^\dagger$~\cite{ECDDP} & Swin-T/7 & No & E-TartanAir & 61.250 & 69.620 & 62.525 & \textbf{74.301} \\ 
    ECDDP$^\ddagger$~\cite{ECDDP} & Swin-T/7 & No & E-TartanAir & 59.142 & 66.450 & - & - \\ 
    ECDDP~\cite{ECDDP} & Swin-T/7 & Yes & E-TartanAir  & 59.820  & 67.680 & 59.748 & 71.056 \\
    \hline
    \algoNameFull (\textbf{Ours}) & Swin-T/7 & Yes & 1Mpx & \textbf{62.774} & \textbf{70.612} & \textbf{65.187} & 72.871 \\  
    \bottomrule
    \end{tabular}
  \vspace{\tablecapmargin}
  \caption{
  Semantic segmentation results on DSEC~\cite{DSEC} and DDD17~\cite{DDD17} datasets.
  Best results are highlighted in \text{bold}.
  We report mean interaction over union (mIoU) and mean class average (mAcc) for evaluation.
  \algoNameFull outperforms state-of-the-art methods in both metrics on DSEC and mIoU for DDD17, and achieves comparable results in mAcc on DDD17.
  $^\dagger$ ECDDP leverages test-time augmentation to improve performance according to their official codebase.
  $^\ddagger$ Our reproduced results by running the official codebase of ECDDP.
  }
  \vspace{\tablemargin}
  \vspace{-3mm}
  \label{tab:sem-seg}
\end{table*}

\vspace{-2mm}
\section{Experiments}
\label{sec:experiments}
\vspace{-1mm}


\subsection{Experimental Setup}

We evaluate \algoNameFull on three downstream tasks: object detection, semantic segmentation, and monocular depth estimation.
Following prior works~\cite{ECDP, ECDDP}, the pre-trained backbone is combined with a task-specific head and fine-tuned on the downstream datasets in a supervised learning manner.
In this subsection, we only introduce the pre-training setting of our method, and leave the implementation details on each downstream task to its own subsection.

\heading{Pre-training dataset.}
We pre-train our backbone on the 1Mpx dataset~\cite{1Mpx} featuring outdoor driving scenarios.
It contains around 15 hours of recordings with a 720$\times$1280 resolution event camera~\cite{1Mpx-camera}, covering both day and night time.
We chose it as 1Mpx has a higher resolution and more moving objects, thus more diverse motions compared to other event camera datasets~\cite{Gen1, DSEC, MVSEC}.

\heading{Recurrent backbone.}
We adopt the Swin-Transformer architecture~\cite{SWIN} with a patch size of 7 (dubbed Swin-T/7) as our encoder $\encoder$ as it is the best-performing backbone in previous event-based SSL methods~\cite{ECDDP}.
To make it a recurrent model, we add ConvLSTM~\cite{ConvLSTM} cells $\recurrent$ after each Swin-T stage.
We take a much smaller version of Swin-T/7 containing only one block as our decoder $\decoder$.
The decoder is a feedforward model without ConvLSTM.
For the input event representation, we follow prior work~\cite{RVT} to aggregate events within every $\intervalLength = 50\, \text{ms}$ into an event histogram, and each event histogram contains $\bins = 10$ temporal bins.
Thanks to the Transformer architecture of Swin-T, our pre-trained backbone can be transferred to any downstream datasets with no resolution limit.

\heading{\algoNameFull pre-training details.}
One training sample contains a sequence of 15 event histograms.
However, this only covers 0.75 seconds, which is still too short in many scenarios.
Therefore, we sequentially sample the same event stream in consecutive training steps.
At each step, we resume from the estimated grayscale video in the previous step and accumulate new input events.
This enables us to train on minute-long event sequences.
We use a masking ratio of 50\%, the decay factor $\alpha = 5$, and a normalizing factor $\denormEvents=$ 5,000.
Yet, as we will see in the ablation study (\cref{sec:ablation}), our model is robust to these hyper-parameters.
The entire model is pre-trained for 400k steps with a batch size of 8.
The Adam optimizer~\cite{Adam} with a peak learning rate of $1 \times 10^{-4}$ and 1.5k warmup steps is used.

\begin{table*}[t]
  \vspace{\pagetopmargin}
  \vspace{-4mm}
  \centering
  \setlength{\tabcolsep}{6.5px}
  \setlength{\arrayrulewidth}{0.5pt} 
  \rowcolors{3}{gray!20}{white}
  \begin{tabular}{lcccccccccc}
    \toprule
    Method & Backbone & Recurrent & Pre-training & {$\delta_1$} $\uparrow$ & {$\delta_2$} $\uparrow$ & {$\delta_3$} $\uparrow$ & Abs $\downarrow$ & RMS $\downarrow$ & RMSlog $\downarrow$ \\
    \midrule
    \multicolumn{3}{l}{\textit{The best performance in the literature.}} \\
    HMNet~\cite{HMNet} & - & Yes & EventScape & 0.588 & 0.784 & 0.889 & 4.171 & 7.534 & 0.397 \\
    \color{gray} PCDepth$^\dagger$~\cite{PCDepth} & \color{gray}- & \color{gray}Yes & \color{gray}EventScape & \color{gray}0.672 & \color{gray}0.845 & \color{gray}0.932 & \color{gray}- & \color{gray}6.621 & \color{gray}0.328 \\ 
    \hline
    No Pre-training & Swin-T/7 & No & - & 0.345 & 0.731 & 0.853 & 5.585 & 8.671 & 0.476 \\  
    RGB Supervised & Swin-T/7 & No & ImageNet-1k & 0.433 & 0.751 & 0.87 & 5.101 & 8.225 & 0.439 \\
    MoBY~\cite{MoBY} & Swin-T/7 & No & ImageNet-1k & 0.5 & 0.753 & 0.868 & 4.974 & 8.066 & 0.433 \\ 
    \hline
    No Pre-training & Swin-T/7 & Yes & - & 0.534 & 0.764 & 0.878 & 4.852 & 7.902 & 0.428 \\  
    RGB Supervised & Swin-T/7 & Yes & ImageNet-1k & 0.596 & 0.799 & 0.906 & 4.116 & 7.174 & 0.375 \\  
    MoBY~\cite{MoBY} & Swin-T/7 & Yes & ImageNet-1k & 0.589 & 0.795 & 0.903 & 4.135 & 7.218 & 0.377 \\  
    \hline
    ECDP~\cite{ECDP} & ResNet50 & No & N-ImageNet &  0.611 & 0.797 & 0.901 & 4.061 & 7.197 & 0.377 \\
    ECDDP~\cite{ECDDP} & ResNet50 & No & E-TartanAir & 0.612 & 0.809 & 0.915 & 3.889 & 6.805 & 0.359 \\ 
    ECDDP~\cite{ECDDP} & Swin-T/7 & No & E-TartanAir & 0.618 & 0.806 & 0.912 & 3.862 & 6.870 & 0.360 \\ 
    ECDDP~\cite{ECDDP} & Swin-T/7 & Yes & E-TartanAir & 0.563 & 0.794 & 0.903 & 4.191 & 7.291 & 0.383\\  
    \hline
    \algoNameFull (\textbf{Ours}) & Swin-T/7 & Yes & 1Mpx & \textbf{0.634} & \textbf{0.830} & \textbf{0.926} & \textbf{3.690} & \textbf{6.654} & \textbf{0.343} \\  
    \bottomrule
  \end{tabular}
  \vspace{\tablecapmargin}
  \caption{
  Monocular depth estimation results on the MVSEC~\cite{MVSEC} dataset, with scores averaged across all testing sequences.
  Best results are highlighted in \text{bold}.
  We report threshold accuracy ($\delta_1$, $\delta_2$, and $\delta_3$), absolute error (Abs), root mean squared error (RMS), and root mean squared logarithmic error (RMSlog) for evaluation.
  $^\dagger$ PCDepth use both images and events as input.
  It only reports results on two sequences and does not release code.
  \algoNameFull outperforms all SSL and event-based methods, and achieve comparable results with PCDepth.
  }
  \vspace{\tablemargin}
  \vspace{-2.5mm}
  \label{tab:depth-est}
\end{table*}

\heading{Baselines.}
We compare our method with two groups of approaches: (i) state-of-the-art methods for each downstream task, and (ii) pre-trained weights from different domains:
\begin{itemize}
    \item \textbf{RGB Supervised:} backbones that are supervised pre-trained on the ImageNet-1k classification dataset~\cite{ImageNet}. We take the official checkpoint from Swin-T/7~\cite{SWIN}.
    \item \textbf{RGB SSL:} we take MoBY~\cite{MoBY} which pre-trains Swin-T/7 on the ImageNet-1k dataset in a self-supervised way.
    We choose it as it is proposed by the authors of Swin-T.
    \item \textbf{Event SSL:} methods that perform unsupervised pre-training of the backbone on the event camera datasets, including ECDP~\cite{ECDP} and the state-of-the-art ECDDP~\cite{ECDDP}.
\end{itemize}
For tasks that are covered by previous works, we simply copy the numbers from their papers.
Otherwise, we take their backbone and fine-tune them using our codebase under the same setting as ours to report results. The detailed setting of pre-training, and downstream task fine-tuning can be found in \cref{sec:experiment-details}. 

\vspace{-1.5mm}
\subsection{Object Detection}
\label{sec:exp-obj-det}
\vspace{-0.75mm}

\heading{Setting.}
We take the detection head design (YOLOX~\cite{yolox}) and the training codebase from a representative recurrent event-based object detector RVT~\cite{RVT}.
We fine-tune all models on Gen1~\cite{Gen1} and 1Mpx~\cite{1Mpx} datasets with the default hyper-parameters in RVT.

\heading{Results.}
\cref{tab:obj-det} presents the quantitative results.
First, we observe that models with recurrent modules consistently outperforms their feedforward counterparts.
This aligns with previous research that shows the importance of history information in the event-based detection task~\cite{RVT, Get-T}.
Compared to no pre-training, pre-training on RGB data improves the performance slightly.
Meanwhile, models pre-trained with \algoNameFull achieve substantial improvement of $3.1\%$ and $5.2\%$ in mAP on Gen1 and 1Mpx, respectively.
This result surpasses ECDDP by $2.0\%$ and $0.6\%$ on both datasets.
Moreover, our model consistently outperforms FCCO~\cite{FCCO} on both datasets, despite FCCO’s use of an optimized event representation and a model architecture 30\% larger than ours (46M vs. 60M parameters).
Overall, our results prove that \algoNameFull unleashes the power of recurrent models to learn long-term information from event sequences, which is crucial for the detection task.

\begin{figure}[t]
    \centering
    \begin{subfigure}{0.32\linewidth}
        \includegraphics[width=1.0\linewidth]{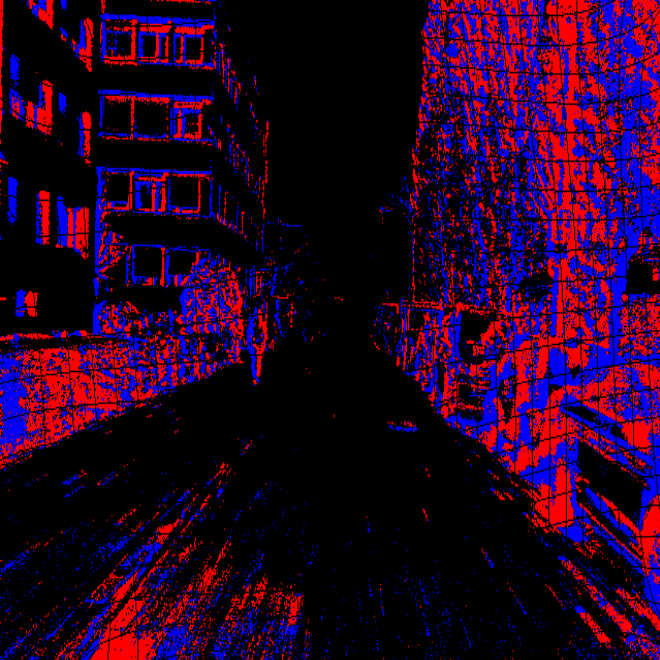}
        \vspace{-4mm}
        \caption{Input events.}
    \end{subfigure}
    \hspace{-0.5mm}
    \begin{subfigure}{0.32\linewidth}
        \includegraphics[width=1.0\linewidth]{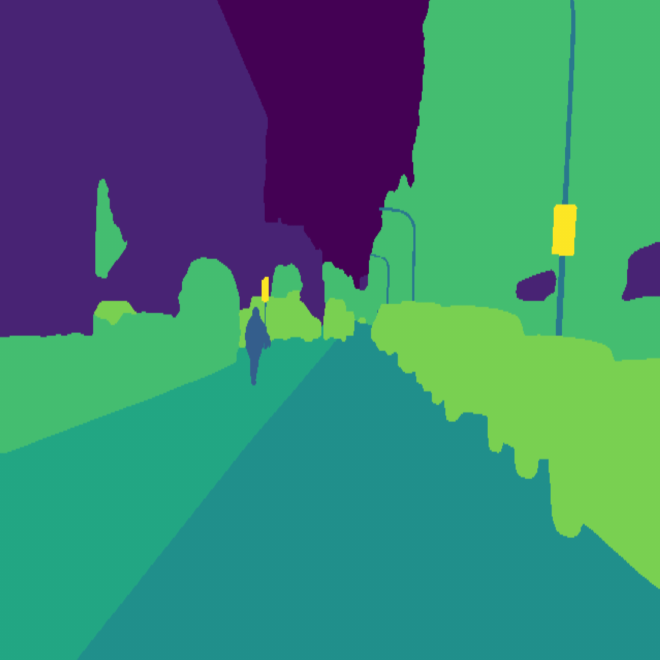}
        \vspace{-4mm}
        \caption{Ground-truth.}
    \end{subfigure}
    \hspace{-0.5mm}
    \begin{subfigure}{0.32\linewidth}
        \includegraphics[width=1.0\linewidth]{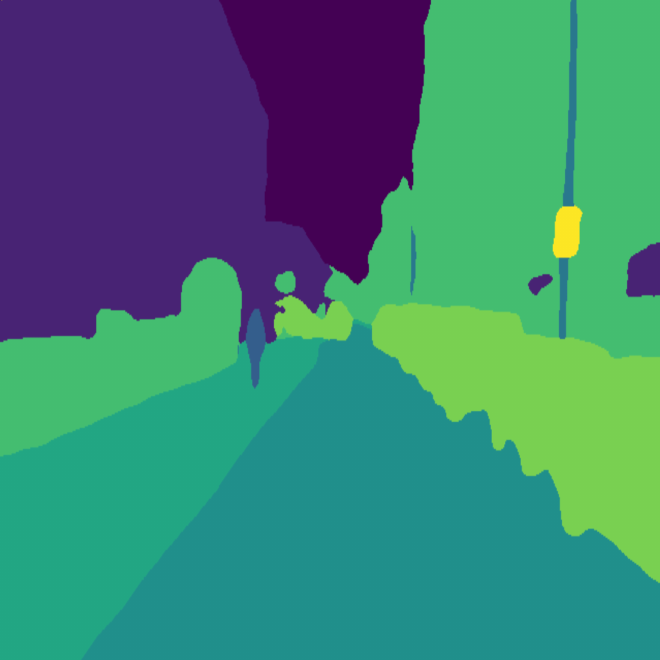}
        \vspace{-4mm}
        \caption{Prediction.}
    \end{subfigure}
    \vspace{\figcapmargin}
    \vspace{-1mm}
    \caption{
        \textbf{Qualitative semantic segmentation results on DSEC.}
        We detect tiny objects such as traffic signs and street lights.
    }
    \label{fig:qual-sem-seg}
    \vspace{\figmargin}
\end{figure}

\begin{figure}[t]
    \vspace{-1mm}
    \centering
    \begin{subfigure}{0.32\linewidth}
        \includegraphics[width=1.0\linewidth]{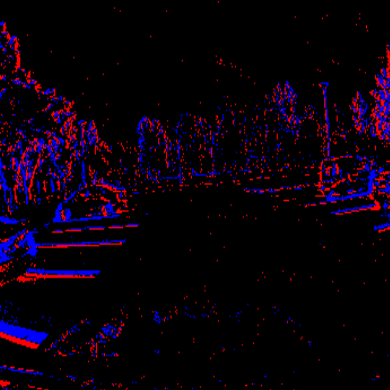}
        \vspace{-4mm}
        \caption{Input events.}
    \end{subfigure}
    \hspace{-0.5mm}
    \begin{subfigure}{0.32\linewidth}
        \includegraphics[width=1.0\linewidth]{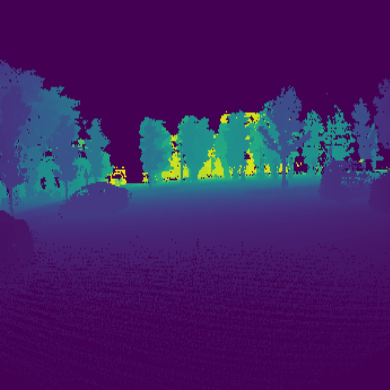}
        \vspace{-4mm}
        \caption{Ground-truth.}
    \end{subfigure}
    \hspace{-0.5mm}
    \begin{subfigure}{0.32\linewidth}
        \includegraphics[width=1.0\linewidth]{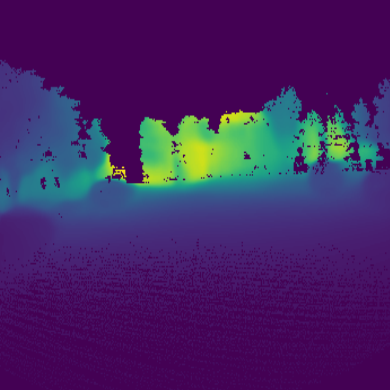}
        \vspace{-4mm}
        \caption{Prediction.}
    \end{subfigure}
    \vspace{\figcapmargin}
    \vspace{-1mm}
    \caption{
        \textbf{Qualitative depth estimation results on MVSEC.}
        We predict accurate depth from sparse events thanks to recurrency.
    }
    \label{fig:qual-depth-est}
    \vspace{\figmargin}
    \vspace{-4.5mm}
\end{figure}

\subsection{Semantic Segmentation}
\label{sec:exp-sem-seg}

\heading{Setting.}
Following prior work~\cite{ECDDP}, we attach an UperNet head~\cite{UperNet} to the backbone and fine-tune all models on the DSEC~\cite{DSEC} and DDD17~\cite{DDD17} datasets.
We train all models for 50k steps with a learning rate of $2 \times 10^{-4}$ and a linear warmup of 250 steps.
Following ESS, the loss function is an unweighted sum of dice loss and cross-entropy loss.
To match the label frequency, we set the event histogram duration $\intervalLength$ to $50 \text{ms}$ for DSEC and $30 \text{ms}$ for DDD17.

\heading{Results.}
\cref{tab:sem-seg} shows the quantitative results.
Recurrent models always outperform their feedfoward counterparts. \algoNameFull surpasses baselines significantly in mIoU on both datasets.
Specifically, we achieve $1.524\%$ higher mIoU on DSEC and $2.275\%$  higher mIoU on DDD17 compared to ECDDP.
For mAcc, we are $0.992\%$ higher on DSEC, and competitive on DDD17.
Notably, ECDDP uses a segmentation head which is $3\times$ larger than ours.
Overall, \algoNameFull brings a gain of $10.499\%$ in mIoU and $9.138\%$ in mAcc on DSEC, and $9.376\%$ in mIoU and $6.105\%$ in mAcc on DDD17 over no pre-training.
This shows our method's effectiveness for pre-training recurrent models.

\begin{table*}
  \vspace{\pagetopmargin}
  \vspace{-5mm}
  \centering
  \begin{tabular}{ccc}
    \begin{subtable}[t]{0.31\textwidth}
        \centering
        \setlength{\tabcolsep}{4.5pt}
        \rowcolors{5}{gray!20}{white}
        \begin{tabular}{ccc}
            \hline
            \multirow{2}{*}{Masking Ratio} & \multicolumn{2}{c}{DSEC} \\ 
            \cmidrule(lr){2-3}
             & mIoU $\uparrow$ & mAcc $\uparrow$ \\
            \hline
            0\% & 55.619 & 62.982 \\
            25\% & 61.625 & 68.715 \\
            \textbf{50\%} & \textbf{62.774} & \textbf{70.612} \\
            75\% & 61.812 & 69.245 \\
            \hline
        \end{tabular}
        \caption{
        \textbf{Masking Ratio.}
        \algoNameFull is robust to masking ratios between $25\%$ and $75\%$, while autoencoding ($0\%$) fails to learn useful representation.
        }
        \label{tab:ablation-mask}
    \end{subtable} &
    \begin{subtable}[t]{0.31\textwidth}
        \centering
        \rowcolors{5}{gray!20}{white}
        \begin{tabular}{ccc}
            \hline
            \multirow{2}{*}{Steps} & \multicolumn{2}{c}{DSEC} \\
            \cmidrule(lr){2-3}
             & mIoU $\uparrow$ & mAcc $\uparrow$ \\
            \hline
            200k & 61.554 & 68.924 \\
            300k & 62.335 & 70.239 \\
            \textbf{400k} & \textbf{62.774} & \textbf{70.612} \\
            500k & 62.336 & 69.993 \\
            \hline
        \end{tabular}
        \caption{
        \textbf{Number of Pre-training Steps.}
        The performance first grows with more training steps, and then saturates at 400k steps.
        }
        \label{tab:ablation-steps}
    \end{subtable} &
    \begin{subtable}[t]{0.31\textwidth}
        \centering
        \rowcolors{5}{gray!20}{white}
        \begin{tabular}{ccc}
            \hline
            \multirow{2}{*}{\denormEvents}  & \multicolumn{2}{c}{DSEC} \\
            \cmidrule(lr){2-3}
             & mIoU $\uparrow$ & mAcc $\uparrow$ \\
            \hline
            500 & 61.742 & 69.072 \\
            \textbf{5,000} & \textbf{62.774} & \textbf{70.612} \\
            50,000 &  61.428 & 68.857 \\
            200,000 & 60.059 & 67.658 \\
            \hline
        \end{tabular}
        \caption{
        \textbf{Normalization Factor in \cref{eq:our-intensity-est}}.
        A value of 5,000 balances the update of new events and the forgetting of old events.
        }
        \label{tab:ablation-norm}
    \end{subtable} \\  

    \begin{subtable}[t]{0.31\textwidth}
        \centering
        \rowcolors{5}{gray!20}{white}
        \begin{tabular}{lcc}
            \toprule
            \multirow{2}{*}{Recon. Target} & \multicolumn{2}{c}{DSEC} \\
            \cmidrule(lr){2-3}
             & mIoU $\uparrow$ & mAcc $\uparrow$ \\
            \midrule
            Histograms & 60.430 & 67.635 \\
            \cref{eq:prior-intensity-est} & 60.493 & 68.021 \\
            \textbf{\cref{eq:our-intensity-est}} & \textbf{62.774} & \textbf{70.612}\\
            \bottomrule
        \end{tabular}
        \caption{
        \textbf{Reconstruction Target.}
        Our improved intensity video (\cref{eq:our-intensity-est}) outperforms both input reconstruction and prior work (\cref{eq:prior-intensity-est}).
        }
        \label{tab:ablation-recon} 
    \end{subtable} &
    \begin{subtable}[t]{0.31\textwidth}
        \centering
        \rowcolors{5}{gray!20}{white}
        \begin{tabular}{lcc}
            \toprule
            \multirow{2}{*}{Dataset} & \multicolumn{2}{c}{DSEC} \\
            \cmidrule(lr){2-3}
             & mIoU $\uparrow$ & mAcc $\uparrow$ \\
            \midrule
            Gen1 & 60.128 & 67.149 \\
            \textbf{1Mpx} & \textbf{62.774} & \textbf{70.612}\\
            \bottomrule
        \end{tabular}
        \caption{
        \textbf{Pre-training Dataset.}
        1Mpx works better than Gen1 as it is of higher resolution, has more objects and diverse motion.
        }
        \label{tab:ablation-dataset}
    \end{subtable} &
    \begin{subtable}[t]{0.31\textwidth}
        \centering
        \setlength{\tabcolsep}{5.5pt}
        \rowcolors{5}{gray!20}{white}
        \begin{tabular}{lcc}
            \toprule
            \multirow{2}{*}{Normalization} & \multicolumn{2}{c}{DSEC} \\
            \cmidrule(lr){2-3}
             & mIoU $\uparrow$ & mAcc $\uparrow$ \\
            \midrule
            w/o norm & 61.792 & 68.705 \\
            \textbf{w/ norm} & \textbf{62.774} & \textbf{70.612}\\ 
            \bottomrule
        \end{tabular} 
        \caption{
        \textbf{Patch Normalization.}
        Computing MSE loss with normalized patches improves the performance, which aligns with prior works~\cite{MAE}.
        }
        \label{tab:ablation-patchnorm}
    \end{subtable}
  \end{tabular}
  \vspace{\tablecapmargin}
  \vspace{-1mm}
  \caption{
  \textbf{Ablation study.}
  Our default settings are \textbf{bold}.
  We report downstream performance on DSEC semantic segmentation.
  }
  \vspace{\tablemargin}
  \vspace{-2mm}
  \label{tab:all-ablation}
\end{table*}
\vspace{-2.5mm}
\subsection{Monocular Depth Estimation}
\label{sec:exp-depth-est}
\vspace{-1.5mm}
\heading{Setting.}
We attach the depth prediction head from MiDaS~\cite{MiDaS} to the backbone and fine-tune all models on the MVSEC~\cite{MVSEC} dataset.
Following prior works~\cite{gehrig2021depth, ECDDP}, we fine-tune on the ``outdoor\_day2" sequence and evaluate on the ``outdoor\_day1", ``outdoor\_night1", ``outdoor\_night2", and ``outdoor\_night3" sequences.
All models are trained for 20k steps with a learning rate of $1 \times 10^{-4}$, a linear warm-up of 100 steps, and a batch size of 8.
We use a weighted combination of a scale-invariant loss and a multi-scale scale-invariant gradient matching loss adopted from the state-of-the-art method HMNet~\cite{HMNet}.

\heading{Results.}
We summarize the quantitative results in \cref{tab:depth-est}.
Due to the high sparsity of the MVSEC dataset, events within a single time segment is not enough for predicting accurate depth.
Therefore, with randomly initialized weights or RGB pre-trained weights, recurrent models consistently outperform feedforward ones.
Similarly, our method designed for recurrent architectures also performs better than ECDP and ECDDP across all metrics by a sizable margin.
Compared to state-of-the-art event-based depth estimators in the literature that are supervised pre-trained on a synthetic dataset~\cite{gehrig2021depth}, \algoNameFull outperforms HMNet~\cite{HMNet}, and achieves comparable results with PCDepth~\cite{PCDepth} which takes in an additional image modality.

\begin{figure}[t]
    \vspace{1mm}
    \centering
    \begin{subfigure}{0.32\linewidth}
        \includegraphics[width=1.0\linewidth]{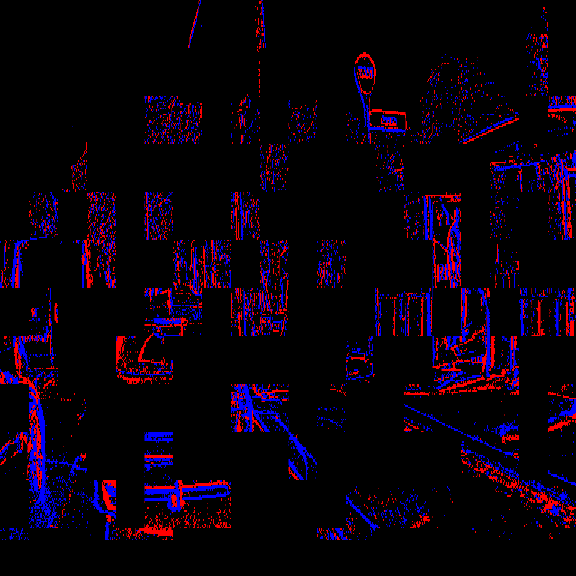}
        \vspace{-4mm}
        \caption{Masked events.}
    \end{subfigure}
    \hspace{-0.5mm}
    \begin{subfigure}{0.32\linewidth}
        \includegraphics[width=1.0\linewidth]{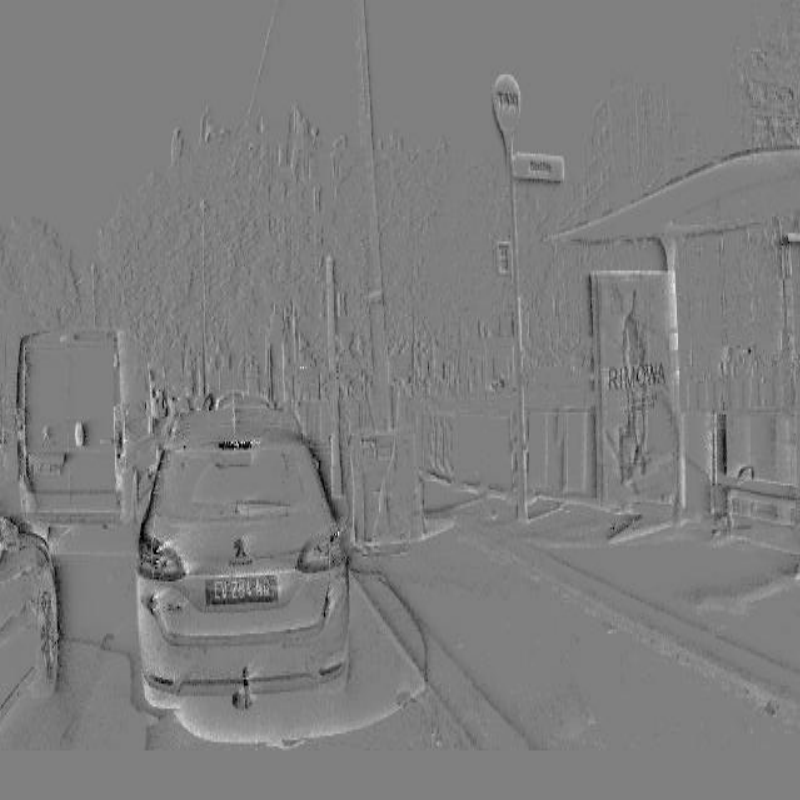}
        \vspace{-4mm}
        \caption{Intensity video.}
    \end{subfigure}
    \hspace{-0.5mm}
    \begin{subfigure}{0.32\linewidth}
        \includegraphics[width=1.0\linewidth]{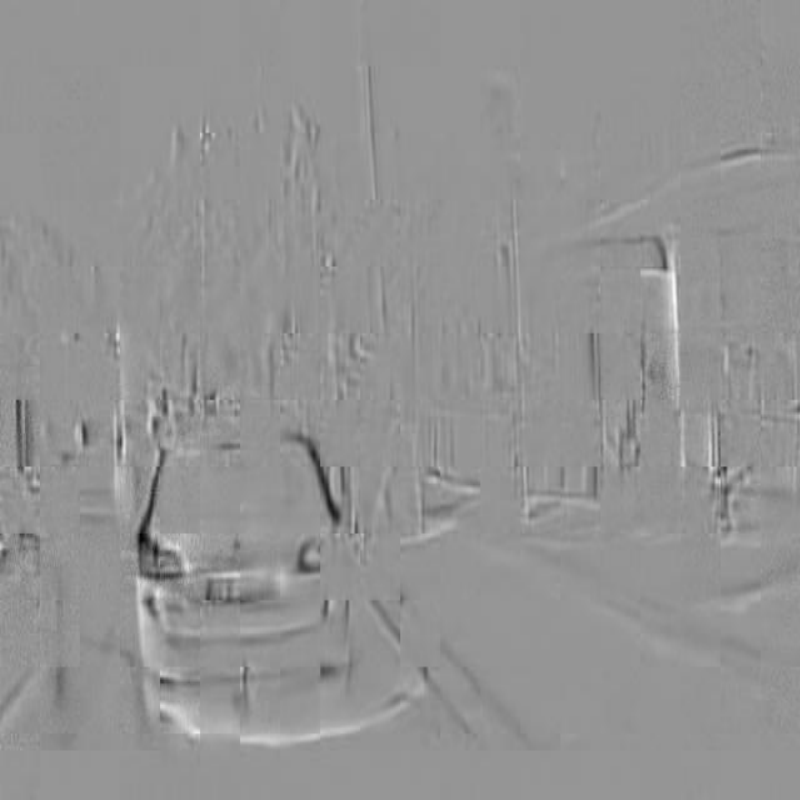}
        \vspace{-4mm}
        \caption{Reconstruction.}
    \end{subfigure}
    \vspace{\figcapmargin}
    \vspace{-1mm}
    \caption{
        \textbf{Qualitative results on masked intensity video reconstruction.}
        Given sparse events that are only partially observable, \algoNameFull produces a holistic reconstruction of the scene.
    }
    \label{fig:qual-recon}
    \vspace{\figmargin}
    \vspace{-5mm}
\end{figure}

\subsection{Ablation Study}
\label{sec:ablation}

We ablate each component of \algoNameFull and report results on the downstream DSEC semantic segmentation task.

\heading{Masking Ratio.}
The masking ratio determines the difficulty of the pre-training task.
\cref{tab:ablation-mask} shows that when not applying masking, \ie doing autoencoding, the pre-training fails to learn useful representation.
When the ratio is low, only a few patches are masked, providing limited signal for training.
On the other hand, a high masking ratio discards too much information, making it challenging to learn.
Given the sparse nature of event data, our optimal value $50\%$ is lower than prior MAE work on RGB videos~\cite{VideoMAENJU}.

\heading{Number of Pre-training Steps.}
\cref{tab:ablation-steps} shows that the performance first scales up with more pre-training steps, and then saturates at 400k steps.

\heading{Normalization Factor $\denormEvents$ in \cref{eq:our-intensity-est}.}
This factor functions as a momentum value, balancing between retaining past information and updating with new inputs.
A small $\denormEvents$ causes the model to forget prior events and only focus on recent inputs, hindering the learning of history information in recurrent models.
A large $\denormEvents$ instead leads to high motion blur, slowing down the model's adaptation to new events.
We chooses $\denormEvents=$ 5,000 which achieves the best balance.

\heading{Reconstruction Target.}
We compare three targets:
(a) input event histograms,
(b) estimated intensity videos from prior work~\cite{Intensity} (\cref{eq:prior-intensity-est}), and
(c) our improved estimated intensity videos (\cref{eq:our-intensity-est}).
\cref{tab:ablation-recon} indicates that reconstructing event histograms leads to worse results.
As discussed in \cref{sec:event-repr}, this is because input events only contain short-term information.
In addition, using our intensity estimation as reconstruction target outperforms the naive one from prior work, highlighting the effectiveness of our modification.
\\
We show a reconstruction results from \algoNameFull in \cref{fig:qual-recon}.
Our model is capable of inpainting partially observable events thanks to the learned long-term history information.

\heading{Pre-training Dataset.}
Previous work~\cite{ECDDP} has shown that a pre-training dataset with diverse scenes and objects can enhance performance.
We draw a similar observation in \cref{tab:ablation-dataset}.
1Mpx provides substantial gains compared to Gen1 due to its higher resolution and more diverse object motions.

\heading{Patch Normalization.}
Similar to MAE on RGB images, \cref{tab:ablation-patchnorm} shows that computing reconstruction loss in \cref{eq:recon-loss} on normalized ground-truth patches improves performance.

\heading{Pre-training Objective.} To assess the effectiveness of MAE, we replace it with alternative contrastive loss functions and report the results in \cref{sec:supp-MAE-justification}.

\vspace{-2mm}
\section{Conclusion}
\vspace{-1mm}
\label{sec:conclusion}
In this paper, we present \algoNameFull, a pure event-based self-supervised pre-training framework.
Our approach adopts the masked image modeling paradigm, and designs a reconstruction target that mines the long-term history information from raw event sequences.
This design greatly enhances the performance of event-based recurrent models.
Equipped with \algoNameFull, recurrent models can beat their feedforward counterparts on multiple event perception tasks.


\vspace{-2mm}
\section*{Acknowledgements}
\vspace{-1mm}
We acknowledge the support of the Natural Sciences and Engineering Research Council of Canada (NSERC). We also acknowledge Vector Institute for computation support.
\clearpage
\appendix
\maketitlesupplementary

\begin{figure*}[t]
    \centering
    \begin{subfigure}{0.238\linewidth}
        \includegraphics[width=1.0\linewidth]{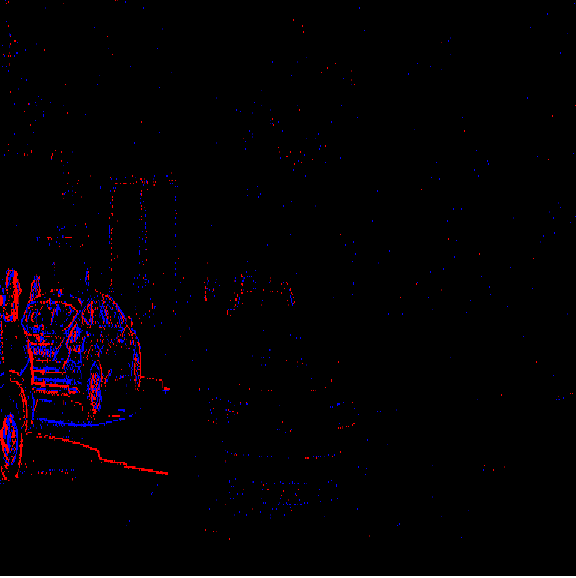}
        \vspace{-4mm}
        \caption{Input events.}
    \end{subfigure}
    \hspace{-0.5mm}
    \begin{subfigure}{0.238\linewidth}
        \includegraphics[width=1.0\linewidth]{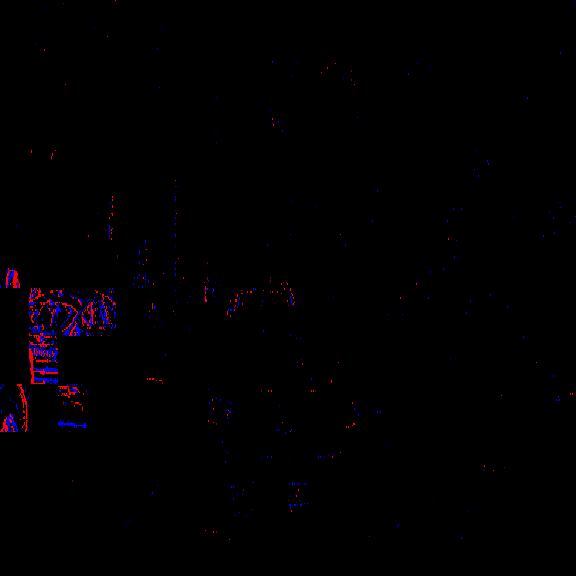}
        \vspace{-4mm}
        \caption{Masked events.}
    \end{subfigure}
    \hspace{-0.5mm}
    \begin{subfigure}{0.238\linewidth}
        \includegraphics[width=1.0\linewidth]{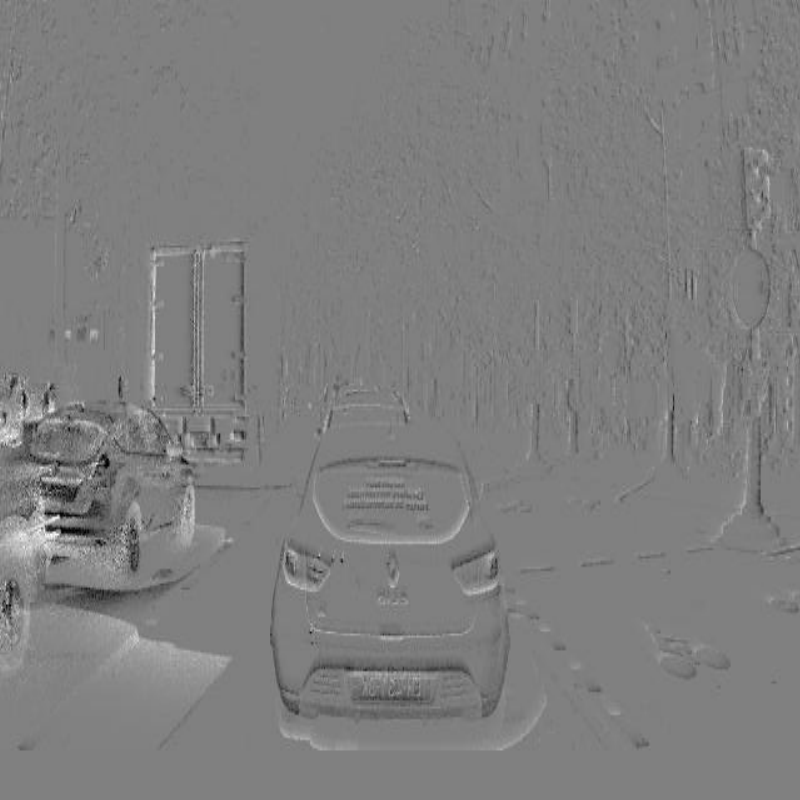}
        \vspace{-4mm}
        \caption{Estimated intensity video.}
    \end{subfigure}
    \hspace{-0.5mm}
    \begin{subfigure}{0.238\linewidth}
        \includegraphics[width=1.0\linewidth]{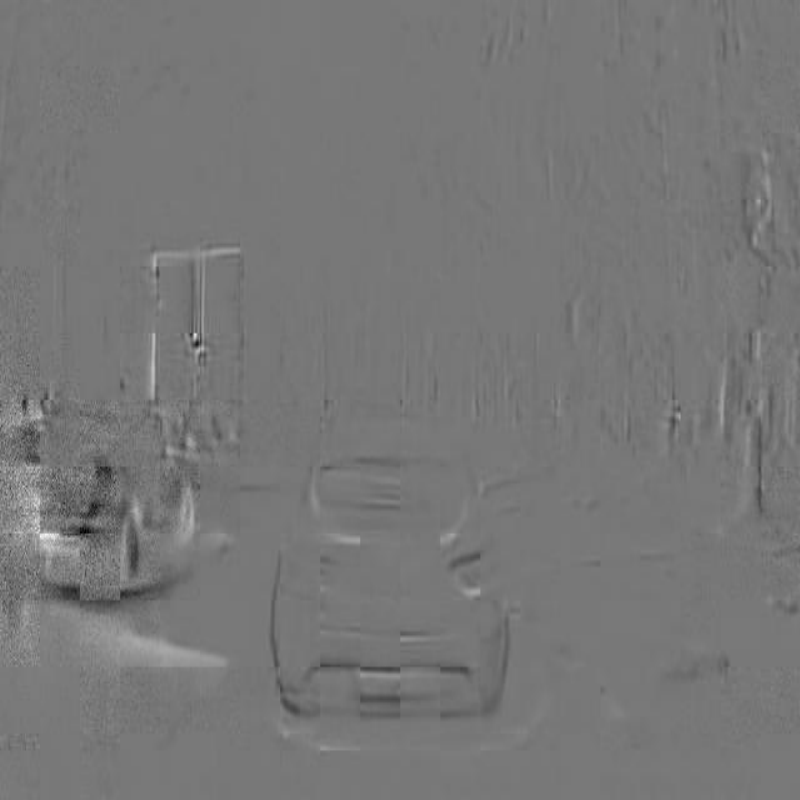}
        \vspace{-4mm}
        \caption{Prediction.}
    \end{subfigure}
    \vspace{\figcapmargin}
    \vspace{-1mm}
    \caption{
        \textbf{Qualitative pre-training results on 1Mpx~\cite{1Mpx}.}
            \algoNameFull is able to reconstruct static objects that are invisible in recent events, \eg the car in front of the ego vehicle.
            This is beneficial to downstream perception tasks such as object detection.
        }
    \label{fig:supp-pretraining-qual}
    \vspace{\figmargin}
\end{figure*}
\begin{figure*}[t]
    \centering
    \begin{subfigure}{0.238\linewidth}
        \includegraphics[width=1.0\linewidth]{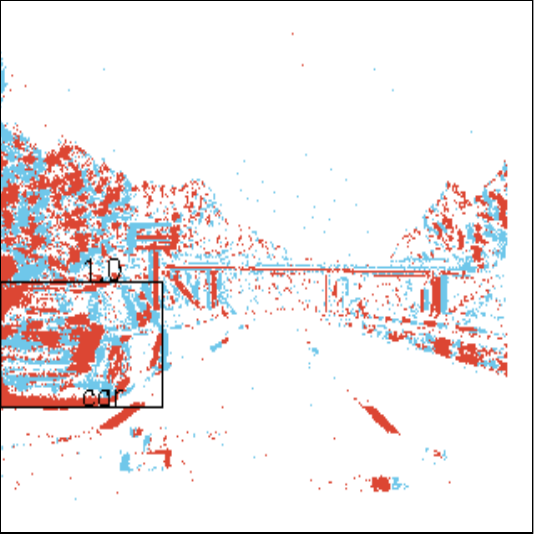}
        \vspace{-4mm}
        \caption{Ground-truth.}
    \end{subfigure}
    \hspace{-0.5mm}
    \begin{subfigure}{0.238\linewidth}
        \includegraphics[width=1.0\linewidth]{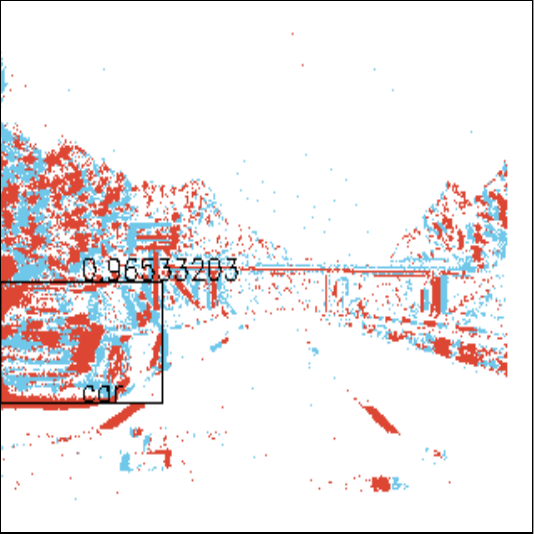}
        \vspace{-4mm}
        \caption{Prediction.}
    \end{subfigure}
    \hspace{-0.5mm}
    \begin{subfigure}{0.238\linewidth}
        \includegraphics[width=1.0\linewidth]{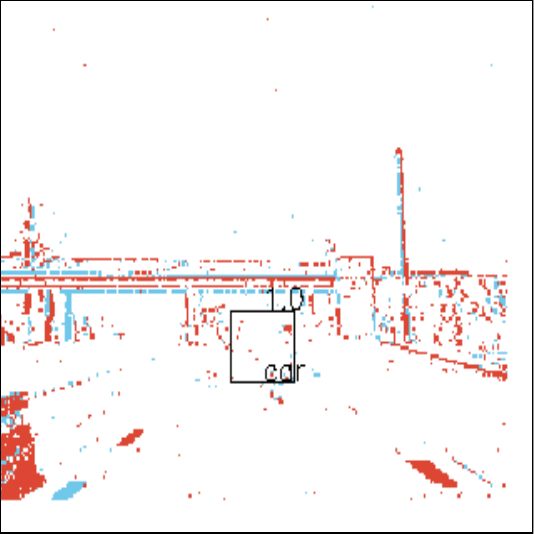}
        \vspace{-4mm}
        \caption{Ground-truth.}
    \end{subfigure}
    \hspace{-0.5mm}
    \begin{subfigure}{0.238\linewidth}
        \includegraphics[width=1.0\linewidth]{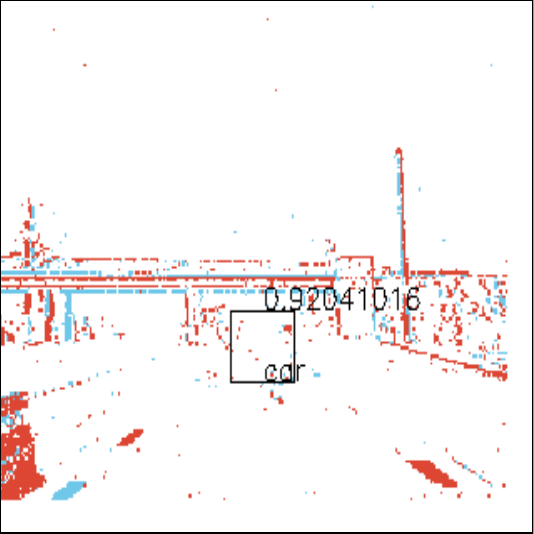}
        \vspace{-4mm}
        \caption{Prediction.}
    \end{subfigure}
    \vspace{\figcapmargin}
    \vspace{-1mm}
    \caption{
        \textbf{Qualitative object detection results on Gen1~\cite{Gen1}.}  
        Thanks to the long-term information learned during the self-supervised pre-training stage, the model successfully detects cars, even when they are not clearly visible in the input data.
    }
    \label{fig:supp-objdet-qual}
    \vspace{\figmargin}
\end{figure*}

\begin{figure*}[t]
    \centering
    \begin{subfigure}{0.32\linewidth}
        \includegraphics[width=1.0\linewidth]{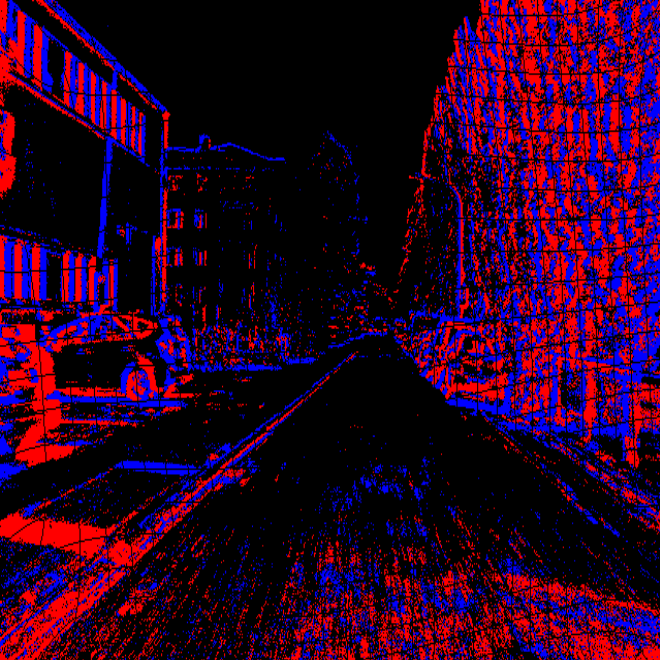}
        \vspace{-4mm}
        \caption{Input events.}
    \end{subfigure}
    \hspace{-0.5mm}
    \begin{subfigure}{0.32\linewidth}
        \includegraphics[width=1.0\linewidth]{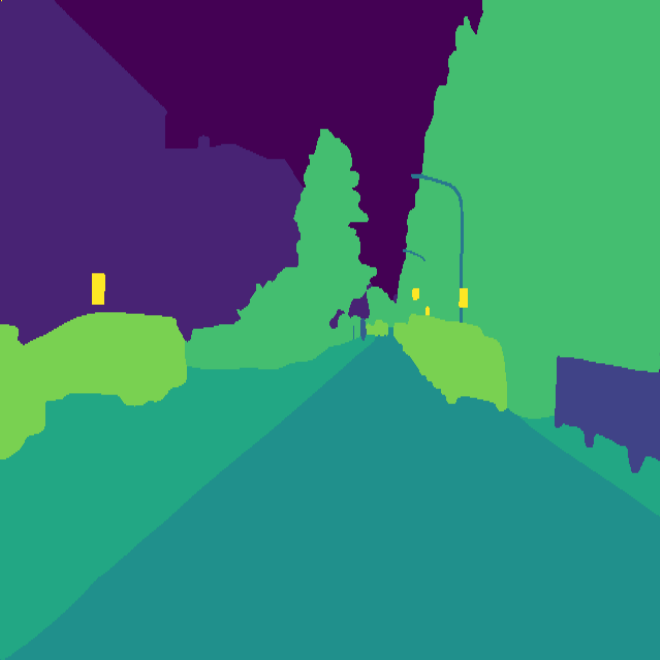}
        \vspace{-4mm}
        \caption{Ground-truth.}
    \end{subfigure}
    \hspace{-0.5mm}
    \begin{subfigure}{0.32\linewidth}
        \includegraphics[width=1.0\linewidth]{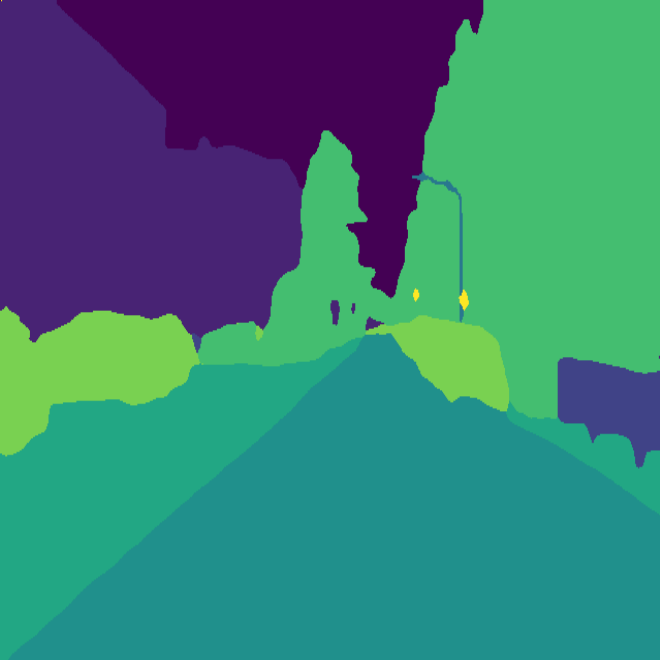}
        \vspace{-4mm}
        \caption{Prediction.}
    \end{subfigure}
    \hspace{-0.5mm}
    \begin{subfigure}{0.32\linewidth}
        \includegraphics[width=1.0\linewidth]{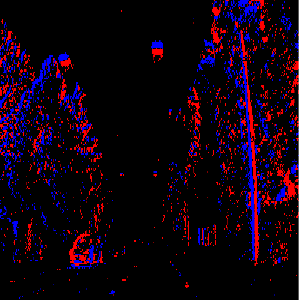}
        \vspace{-4mm}
        \caption{Input events.}
    \end{subfigure}
    \hspace{-0.5mm}
    \begin{subfigure}{0.32\linewidth}
        \includegraphics[width=1.0\linewidth]{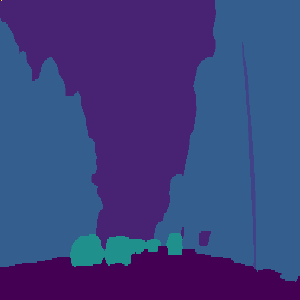}
        \vspace{-4mm}
        \caption{Ground-truth.}
    \end{subfigure}
    \hspace{-0.5mm}
    \begin{subfigure}{0.32\linewidth}
        \includegraphics[width=1.0\linewidth]{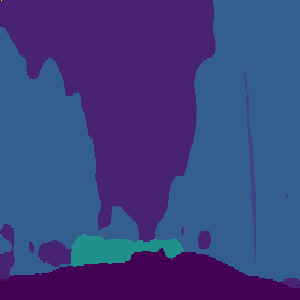}
        \vspace{-4mm}
        \caption{Prediction.}
    \end{subfigure}
    \hspace{-0.5mm}
    \vspace{\figcapmargin}
    \vspace{-1mm}
    \caption{
        \textbf{Qualitative semantic segmentation results on DSEC~\cite{DSEC} and DDD17~\cite{DDD17}.}  
        Figures (a-c) are from the DSEC dataset, and figures (d-f) are from the DDD17 dataset.
        Leveraging long-term information learned in the pre-training stage, the fine-tuned models achieve high accuracy in predicting segmentation maps across datasets with varying resolutions and sparse event frames.
    }
    \label{fig:supp-semseg-qual}
    \vspace{\figmargin}
\end{figure*}

\begin{figure*}[t]
    \centering
    \begin{subfigure}{0.32\linewidth}
        \includegraphics[width=1.0\linewidth]{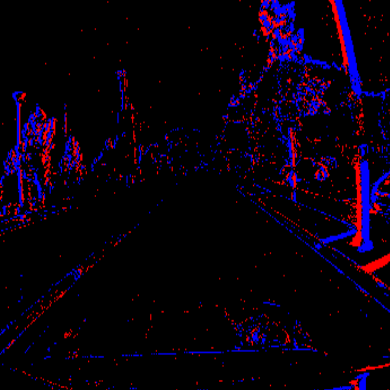}
        \vspace{-4mm}
        \caption{Input events.}
    \end{subfigure}
    \hspace{-0.5mm}
    \begin{subfigure}{0.32\linewidth}
        \includegraphics[width=1.0\linewidth]{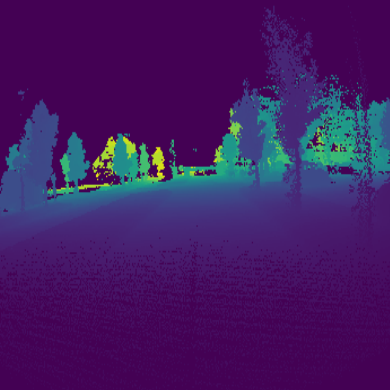}
        \vspace{-4mm}
        \caption{Ground-truth.}
    \end{subfigure}
    \hspace{-0.5mm}
    \begin{subfigure}{0.32\linewidth}
        \includegraphics[width=1.0\linewidth]{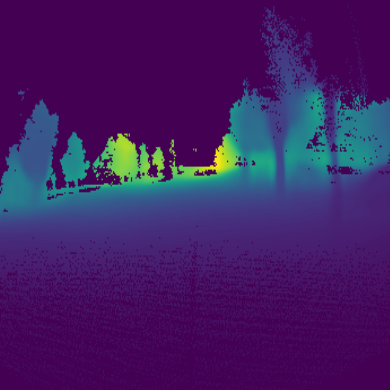}
        \vspace{-4mm}
        \caption{Prediction.}
    \end{subfigure}
    \vspace{\figcapmargin}
    \vspace{-1mm}
    \caption{
        \textbf{Qualitative monocular depth estimation results on MVSEC~\cite{MVSEC}.}  
        Thanks to the long-term information learned with \algoNameFull, the model detects the surfaces of various objects and predicts accurate depth maps from sparse event frames.
    }
    \label{fig:supp-depest-qual}
    \vspace{\figmargin}
\end{figure*}

\section{Additional Experimental Results}
\label{sec:supp-additional-results}
In this section, we provide further analysis to complement the experimental results presented in the main paper. 
This includes an expanded ablation study to show the effectiveness of our improved intensity estimation video, qualitative results that highlight the strengths of our approach through visual examples, and a discussion of the trade-off involved in choosing the temporal bin  in \cref{eq:our-intensity-est}.

\subsection{Ablation Study}
\label{sec:supp-ablation}
We present additional evidence to show the effectiveness of our improved intensity video reconstruction method, \cref{eq:our-intensity-est}, compared to the naive approach from prior work, \cref{eq:prior-intensity-est}.
The performance of fine-tuned models on the downstream DDD17~\cite{DDD17} semantic segmentation task is summarized in \cref{tab:supp-ablation}. 
Using our intensity estimation as the reconstruction target achieves improvements of $1.352\%$ in mIoU and $2.691\%$ in mAcc over the naive estimation.

\subsection{Qualitative Results}
\label{sec:supp-qual}
We provide additional qualitative results to demonstrate the performance of our model in both self-supervised pre-training and downstream perception tasks.

\heading{Pre-training.}  
The visualized sample in \cref{fig:supp-pretraining-qual} shows \algoNameFull reconstruction of the scene. 
This representation includes static objects that were invisible in recent event frames. 
This example highlights that the recurrent design of \algoNameFull helps it to extract a rich representation of various objects in the scene, beyond those that are actively moving.

\heading{Object Detection.}  
Qualitative samples of object detection on the Gen1~\cite{Gen1} dataset are presented in \cref{fig:supp-objdet-qual}. 
The model successfully detects cars in the scene, even when they are barely visible in recent events.

\heading{Semantic Segmentation.}  
\cref{fig:supp-semseg-qual} showcases examples from the DSEC~\cite{DSEC} and DDD17~\cite{DDD17} datasets. 
These results show that our pre-trained model adapt effectively to downstream datasets with varying resolutions. 
Notably, the model predicts accurate segmentation maps, even with sparse inputs.

\heading{Monocular Depth Estimation.}  
Qualitative results in \cref{fig:supp-depest-qual} show that the model accurately distinguishes between different objects and predicts precise depth maps.

\subsection{Choosing the temporal bin in Eq. (7)}
While using small temporal bins increases temporal resolution, it also increases the computational cost, as more iterations are required to update the reconstruction target. 
In addition, noises, such as hot pixels, might dominate some of the pixels. 
On the other hand, large bins result in severe information loss due to a large $\Delta$.
We find that a 5ms bin size balances target precision and noise robustness.

\section{Justification of MAE for sparse event data} 
\label{sec:supp-MAE-justification}
Although the event input is sparse, \algoNameFull reconstructs a \emph{dense pseudo gray-scale} video as target. Therefore, reconstructing masked patches of our target provides a strong training signal.
Moreover, masking a large portion of the event input along the time requires the model to \emph{estimate spatiotemporal information}.
This helps the model's scene understanding and leads to higher performance compared to training with no masking (see \cref{tab:ablation-mask}). 
To assess the impact of the pre-training objective, we replace the MAE~\cite{VideoMAEV2} loss in \algoNameFull with two alternative loss functions: 
(i) Contrastive Predictive Coding (CPC)~\cite{CPC,CPCV2}, which learns representations by predicting future latent features using contrastive loss, and (ii) the contrastive loss introduced in ECDDP~\cite{ECDDP}. Their downstream performance on the semantic segmentation task on the DSEC dataset is reported in \cref{tab:supp-ablation-objective}. 
While both contrastive losses lead to improved performance over the baseline recurrent architecture without pre-training, MAE yields the highest accuracy. 
This demonstrates the effectiveness of MAE as a pre-training objective in \algoNameFull.

\section{Implementation Details}
\label{sec:experiment-details}

We describe the implementation settings used for \algoNameFull and downstream tasks, including object detection, semantic segmentation, and monocular depth estimation. 
\cref{tab:all-config} summarizes the settings and hyper-parameters.

\heading{Data Pre-processing.}
We adopt a unified event representation across \algoNameFull and all downstream tasks. 
An event stream is split into non-overlapping event segments of length $\intervalLength$. 
Each segment is then converted into a 2D histogram with 10 bins for positive events and 10 bins for negative ones. 
Input values are clipped between 0 and 10 to prevent the influence of hot pixels. 
The input resolution is required to be divisible by 32 due to the characteristics of the used architecture. 
For datasets with resolutions that do not meet this requirement, zero-padding is applied to align the input dimensions. 
Data augmentation is performed using flipping and scaling transformations. The probabilities and parameters of transformations are detailed in \cref{tab:all-augmentation}.

\begin{table}[t]
    \rowcolors{2}{white}{gray!20}
    \setlength{\tabcolsep}{19pt}
    \centering
    \begin{tabular}{lcc}
        \toprule
        \multirow{2}{*}{Recon. Target} & \multicolumn{2}{c}{DDD17} \\
        \cmidrule(lr){2-3}
         & mIoU $\uparrow$ & mAcc $\uparrow$ \\
        \midrule
        \cref{eq:prior-intensity-est} & 63.835 & 70.180 \\
        \textbf{\cref{eq:our-intensity-est}} & \textbf{65.187} & \textbf{72.871}\\
        \bottomrule
    \end{tabular}
    \vspace{\tablecapmargin}
    \caption{
    \textbf{Ablation on the reconstruction target.}
    We report downstream performance on DDD17~\cite{DDD17} semantic segmentation.
    }
    \label{tab:supp-ablation}
\end{table}

\begin{table}[t]
    \rowcolors{2}{white}{gray!20}
    \setlength{\tabcolsep}{14.2pt}
    \centering
    \begin{tabular}{lcc}
        \toprule
        \multirow{2}{*}{Pre-training Objective} & \multicolumn{2}{c}{DSEC} \\
        \cmidrule(lr){2-3}
         & mIoU $\uparrow$ & mAcc $\uparrow$ \\
        \midrule
        CPC~\cite{CPC} & 59.921 & 67.486\\
        ECDDP~\cite{ECDDP} & 61.012 & 68.231\\
        MAE~\cite{MAE} & \textbf{62.774} & \textbf{70.612} \\
        \bottomrule
    \end{tabular}
    \vspace{\tablecapmargin}
    \caption{
    \textbf{Ablation on the pre-training objective.}
    We report downstream performance on DSEC~\cite{DSEC} semantic segmentation.
    }
    \vspace{\tablemargin}
    \vspace{-1mm}
    \label{tab:supp-ablation-objective}
\end{table}

\heading{Dataloading.}
During the training phase of recurrent models, we process multiple stages within a single training iteration. 
The total number of stages is determined by the sequence length.
To enable the model to handle long sequences in \algoNameFull and object detection, we adopt the dataloading mechanism from RVT~\cite{RVT}. 
This approach allows the model to be trained on minute-long sequences.

\heading{Recurrent Backbone.}  
Following recent work~\cite{ECDDP}, we choose the Swin Transformer~\cite{SWIN} architecture with a window size of 7 (referred to as Swin-T/7) as the encoder for both pre-training and downstream tasks.
The implementation of this architecture is borrowed from Timm~\cite{timm}. 
Since we employ a 2D histogram representation with 20 bins for events, the first embedding layer of Swin-T/7 is modified to accept 20 input channels instead of 3. 
For baselines where pre-trained weights with a different representation are used, the embedding layer's weights are repeated to match the 20 channels. 
A ConvLSTM~\cite{ConvLSTM} layer is inserted after each Swin Block to add recurrency into the backbone.
The number of parameters for backbones and task-specific heads used in this work are show in \cref{tab:supp-modulesize}.

\heading{Optimization.}
We use the Adam~\cite{Adam} optimizer for pre-training and all tasks, along with a learning rate scheduler.
The schedulers, learning rates, and warm-up steps are specified in \cref{tab:all-config}. 
The initial and final learning rates are set to the peak learning rate divided by $c_1$ and $c_2$, respectively.

\heading{Evaluation.}
For evaluating feedforward models during test time, an event segment is extracted for each label from the corresponding event stream. 
As a result, the inputs for consecutive labels may overlap. 
In contrast, recurrent models can process the events of an entire stream only once and predict all labels for that stream.
This approach requires setting $\intervalLength$ based on the label frequency for each dataset in downstream tasks. 
Since the time intervals between consecutive labels may vary slightly, all events occurring between two labels are used as the input for the later label.

\begin{table}
    \begin{subtable}[t]{\columnwidth}
        \centering
        \setlength{\tabcolsep}{9pt}
        \rowcolors{2}{white}{gray!20}
        \begin{tabular}{l|c}
            \hline
            Module & Parameters (M) \\ 
            \hline
            Feedforward backbone & 27.5 \\
            Recurrent backbone & 33.8 \\
            \hline
            \algoNameFull decoder head & 3.7 \\
            Object Detection head & 12.9 \\
            Semantic Segmentation head & 11.2 \\
            Monocular Depth Estimation head & 4.1\\
            \hline            
        \end{tabular}
    \end{subtable}
    \vspace{\tablecapmargin}
    \caption{
        Size of backbones and task-specific heads.
    }
    \vspace{\tablemargin}
    \label{tab:supp-modulesize}
\end{table} 

\begin{table*}
\centering
\begin{tabular}{cc}
    \begin{subtable}[t]{0.4\textwidth}
        \centering
        \setlength{\tabcolsep}{16pt}
        \rowcolors{2}{white}{gray!20}
        \begin{tabular}{l|c}
            \hline
            Architecture & Recurrent \\ 
            Dataset & 1Mpx \\ 
            \hline
            Effective Batch Size & 8 \\
            Peak Learning Rate & 2e-4\\
            $c_1$, $c_2$ & 25, 1,000\\
            Optimizer & Adam\\
            Scheduler & Linear \\ 
            Training Steps & 400,000 \\
            Warm Up Steps &  2,000 \\
            $\intervalLength$ & 50ms \\
            Sequence Length & 15\\
            Masking Ratio & $50\%$ \\
            $\denormEvents$ & 5,000 \\
            GPU & $4\times$ RTX6000 \\
            \hline
        \end{tabular}
        \caption{
        \algoNameFull pre-training stage
        }
        \label{tab:config-tessl}
    \end{subtable} &
    \begin{subtable}[t]{0.55\textwidth}
        \centering
        \setlength{\tabcolsep}{8pt}
        \renewcommand{\arraystretch}{1.27} 
        \rowcolors{2}{white}{gray!20}
        \begin{tabular}{l|cccc}
            \hline
            Architecture & \multicolumn{2}{c}{Recurrent} & \multicolumn{2}{c}{Feedforward} \\ 
            Dataset & Gen1 & 1Mpx & Gen1 & 1Mpx \\ 
            \hline
            Effective Batch Size & 8 & 24 & 64 & 32 \\
            Peak Learning Rate & 2e-4 & 3.46e-4 & 2e-4 & 3.46e-4\\
            $c_1$, $c_2$ & \multicolumn{4}{c}{25, 1,000}\\
            Optimizer & \multicolumn{4}{c}{Adam}\\
            Scheduler & \multicolumn{4}{c}{Linear} \\ 
            Training Steps & \multicolumn{4}{c}{400,000} \\
            Warm Up Steps &  \multicolumn{4}{c}{2,000} \\
            $\intervalLength$ & \multicolumn{4}{c}{50ms} \\
            Sequence Length & 21 & 5 & \multicolumn{2}{c}{1}\\
            GPU & \multicolumn{2}{c}{$2\times$ RTX6000} & \multicolumn{2}{c}{$1\times$ RTX6000} \\
            \hline
        \end{tabular}
        \caption{
        Object Detection
        }
        \label{tab:config-objdet}
    \vspace{2mm}
    \end{subtable}
    \\
    \begin{subtable}[t]{0.4\textwidth}
        \centering
        \setlength{\tabcolsep}{5pt}
        \rowcolors{2}{white}{gray!20}
        \begin{tabular}{l|cc}
            \hline
            Architecture & Recurrent & Feedforward \\ 
            Dataset & \multicolumn{2}{c}{MVSEC}\\ 
            \hline
            Effective Batch Size & 8 & 16 \\
            Peak Learning Rate & \multicolumn{2}{c}{1e-4}\\
            $c_1$, $c_2$ & \multicolumn{2}{c}{25, 1,000}\\
            Optimizer & \multicolumn{2}{c}{Adam}\\
            Scheduler & \multicolumn{2}{c}{Cosine} \\ 
            Training Steps & \multicolumn{2}{c}{20,000} \\
            Warm Up Steps &  \multicolumn{2}{c}{100} \\
            $\intervalLength$ & 50ms & 100ms \\
            Sequence Length & 10 & 1\\
            $d_{\text{max}}, \alpha$ &  \multicolumn{2}{c}{80, 3.7}  \\
            GPU & \multicolumn{2}{c}{$1\times$ RTX6000} \\
            \hline
        \end{tabular}
        \caption{
        Monocular Depth Estimation
        }
        \label{tab:config-depth}
    \end{subtable} &
    \begin{subtable}[t]{0.55 \textwidth}
        \centering
        \setlength{\tabcolsep}{7pt}
        \renewcommand{\arraystretch}{1.17} 

        \rowcolors{2}{white}{gray!20}
        \begin{tabular}{l|cccc}
            \hline
            Architecture & \multicolumn{2}{c}{Recurrent} & \multicolumn{2}{c}{Feedforward} \\ 
            Dataset & DSEC & DDD17 & DSEC & DDD17 \\ 
            \hline
            Effective Batch Size & 8 & 16 & \multicolumn{2}{c}{16} \\
            Peak Learning Rate & \multicolumn{4}{c}{1e-4}\\
            $c_1$, $c_2$ & \multicolumn{4}{c}{25, 1,000}\\
            Optimizer & \multicolumn{4}{c}{Adam}\\
            Scheduler & \multicolumn{4}{c}{Cosine} \\ 
            Training Steps & \multicolumn{4}{c}{50,000} \\
            Warm Up Steps &  \multicolumn{4}{c}{250} \\
            $\intervalLength$ & 50ms & 30ms & \multicolumn{2}{c}{100ms} \\
            Sequence Length & 10 & 15 & \multicolumn{2}{c}{1}\\
            GPU & \multicolumn{2}{c}{$2\times$ RTX6000} & \multicolumn{2}{c}{$1\times$ RTX6000} \\
            \hline
        \end{tabular}
        \caption{
        Semantic Segmentation
        }
        \label{tab:config-semseg}
    \end{subtable}\\
\end{tabular}
\caption{
\textbf{Implementation Settings.}  
We report the implementation details for both pre-training and downstream tasks.
}
\label{tab:all-config}
\end{table*}

\begin{table}
    \begin{subtable}[t]{\columnwidth}
        \centering
        \setlength{\tabcolsep}{12pt}
        \rowcolors{2}{white}{gray!20}
        \begin{tabular}{lccc}
            \hline
            \multirow{2}{*}{Augmentation} & \multirow{2}{*}{Probability} & \multicolumn{2}{c}{Magnitude} \\ 
             \cmidrule(lr){3-4} 
             &  & min & max \\ 
            \hline
            Horizontal Flip & 0.5 & - & - \\
            Apply Zoom & 0.8 & - & - \\
            Zoom In & 0.8 & 1 & 1.5 \\
            Zoom Out & 0.2 & 1 & 1.2 \\ 
            \hline
        \end{tabular}
        \caption{
            \algoNameFull, Object Detection, and Semantic Segmentation.
        }
        \label{tab:augmentation-double}
    \end{subtable}
    \\[3mm]
    \begin{subtable}[t]{\columnwidth}
        \centering
        \setlength{\tabcolsep}{32pt}
        \rowcolors{2}{white}{gray!20}
        \begin{tabular}{lc}
            \hline
            Augmentation & Probability \\ 
             \hline
            Horizontal Flip & 0.5 \\
            \hline
        \end{tabular}
        \caption{
            Monocular Depth Estimation.
        }
        \label{tab:augmentation-single}
    \end{subtable}
    \caption{
        \textbf{Augmentation Transformations.}  
        Description of data augmentations used during training.
    }
    \label{tab:all-augmentation}
\end{table}

\subsection{\algoNameFull Pre-training}
\label{sec:config-pretraining}
Our pre-training hyper-parameters are detailed in \cref{tab:config-tessl}.
\heading{Dataset.}
We use the 1Mpx~\cite{1Mpx} dataset as the default for pre-training. 
This dataset contains approximately 15 hours of autonomous driving scenarios, captured during both daytime and nighttime. 
The original resolution of the dataset is $720 \times 1280$, but we downsample it by 2  to reduce computational costs. 
Compared to other event camera datasets, 1Mpx features a higher resolution and a greater number of moving objects, resulting in more diverse motion patterns.

\heading{Data Representation.}
We set $\intervalLength$ to 50ms for pre-training.

\heading{Masking.} 
To perform the masking, each bin of the histograms is divided into $32 \times 32$ non-overlapping patches. 
Then, $50\%$ of the patches are randomly masked. 
The masked patches are replaced with a $32 \times 32$ learnable parameter. 
The same mask is applied across all bins within a single training step following VideoMAE~\cite{VideoMAEV2}.

\heading{Architecture.} 
We adopt an asymmetric design for the encoder-decoder paradigm, where the decoder consists of a single Swin Block with a window size of 7. 
The Swin block is followed by a convolutional layer to match the reconstructed output size with the input size.

\heading{Loss Function.}
We apply the Mean Squared Error (MSE) loss only to the masked patches. 
The loss value is computed between the predictions and the normalized ground-truth patches following prior works~\cite{MAE, VideoMAENJU}.

\subsection{Object Detection}
We fine-tune our models on the Gen1~\cite{Gen1} and 1Mpx~\cite{1Mpx} datasets for object detection as one of the downstream tasks.

\heading{Datasets.}
The characteristics of the 1Mpx dataset are described in \cref{sec:config-pretraining}. 
Gen1 dataset is a dataset for detecting objects from event cameras mounted on vehicles. 
It contains 2,358 event sequences, each lasting 60 seconds (39 hours in total) with a resolution of $304 \times 240$ pixels.

\heading{Data Representation.}
We set $\intervalLength$ to 50ms for all models following RVT~\cite{RVT}.

\heading{Architecture.}
We adopt the RVT architecture design for object detection, with one key difference: we replace their recurrent backbone with our Swin-T/7 recurrent backbone. Specifically, we use the YOLOX framework~\cite{yolox}, which includes intersection over union (IoU) loss, classification loss, and regression loss. These losses are averaged over both the batch and sequence length for each optimization step.

\heading{Training.}
We fine-tune our models for 400,000 training steps. 
Experimentally, we found that a learning rate of $1 \times 10^{-4}$ achieves the best performance on the 1Mpx dataset when pre-training is conducted on the same dataset. 
We hypothesize that this is because the pre-trained model already extracts informative representations of the data, which reduces the need for a larger learning rate during fine-tuning. 
All other hyper-parameters are detailed in \cref{tab:config-objdet}.

\subsection{Semantic Segmentation}
We fine-tune our models on the DSEC~\cite{DSEC} and DDD17~\cite{DDD17} datasets for semantic segmentation.

\heading{Datasets.}
DSEC and DDD17 are both autonomous driving datasets. The DSEC dataset has a resolution of $640 \times 480$ and contains 53 sequences. However, semantic maps are available for only 11 of them. The DDD17 dataset is relatively longer, with a resolution of $346 \times 260$ and 40 sequences. Similar to DSEC, semantic maps for DDD17 are provided for only 6 sequences.

\heading{Data Representation.}
DSEC and DDD17 labels are available at 20\,Hz and 33\,Hz, respectively. Consequently, $\intervalLength$ is set to 50\,ms for DSEC and 30\,ms for DDD17 when using recurrent models. For feedforward baselines, $\intervalLength$ is set to 100\,ms to avoid extremely sparse inputs.

\heading{Architecture.}
We fine-tune the backbones with an attached UperNet~\cite{UperNet} decoder.

\heading{Loss Function.}
We use the sum of cross-entropy and Dice loss~\cite{DiceLoss} as the loss function, as suggested in~\cite{ESS}.

\subsection{Monocular Depth Estimation}
We fine-tune our models on MVSEC~\cite{MVSEC} dataset for the monocular depth estimation task.

\heading{Dataset.}
The MVSEC dataset consists of event data and grayscale images recorded by a DAVIS event camera with a resolution of $346 \times 260$ pixels, mounted on a driving car. 
Ground-truth depth maps are recorded at 20\,Hz by a LiDAR sensor. 
The dataset contains several sequences captured during daytime and nighttime. 
Following ECDDP~\cite{ECDDP}, we use the ``outdoor\_day2'' sequence for fine-tuning and the ``outdoor\_day1'', ``outdoor\_night1'', ``outdoor\_night2'', and ``outdoor\_night3'' sequences for evaluation.

\heading{Data Representation.}
$\intervalLength$ is set to 50\,ms for recurrent models, based on the MVSEC label frequency (20\,Hz). For feedforward models, $\intervalLength$ is set to 100\,ms.

\heading{Architecture.}
We attach a depth prediction head from MiDaS~\cite{MiDaS} to our backbone.

\heading{Loss Function.}
Following HMNet~\cite{HMNet}, we train the model to predict the normalized log depth $\hat{d}$, defined as:
\begin{align}
\label{eq:depth-loss}
    \hat{d} = \frac{1}{\alpha} \log\frac{d}{d_{\text{max}}} + 1,
\end{align}
where $d$ is the metric depth, $d_{\text{max}}$ is the maximum depth in the dataset, and $\alpha$ is determined by the ratio between the maximum depth $d_{\text{max}}$ and the minimum depth $d_{\text{min}}$:
\begin{align}
\label{eq:alpha}
    \alpha = \log\frac{d_{\text{max}}}{d_{\text{min}}}.
\end{align}
For the MVSEC dataset, $d_{\text{max}}$ and $\alpha$ equal to 80 and 3.7.

We train our models using the same loss function as in previous work~\cite{gehrig2021depth}. 
We compute a weighted sum of the scale-invariant loss~\cite{SIL} and the multi-scale scale-invariant gradient matching loss~\cite{GML} as the loss function. 
The weights for the scale-invariant loss and the gradient matching loss are set to 1 and 0.125, respectively.

\begin{table}[t]
    \centering
    \renewcommand{\arraystretch}{0.9}
    \setlength{\tabcolsep}{13pt}
        \begin{tabular}{c c c c}
            \toprule
             & ECDP & ECDDP & \algoNameFull \\           
            \midrule
            \rowcolor{gray!20} Runtime & 45h & 120h & 135h \\
            \bottomrule
        \end{tabular}%
    \vspace{\tablecapmargin}
    \caption{
        \textbf{Runtime comparison of event-based SSL methods.} Each experiment is conducted on four RTX6000 GPUs.
    }
    \label{tab:supp-pretrain-runtime}
    \vspace{\tablemargin}
\end{table}

\begin{table}[t]
    \centering
    \setlength{\tabcolsep}{6.6pt}
    \footnotesize
  \begin{tabular}{l c c c c c}
    \toprule
    \multirow{2}{*}{Recurrent} & \multicolumn{2}{c}{Semantic Seg.} & Depth Est. & \multicolumn{2}{c}{Object Det.} \\ 
    \cmidrule(lr){2-3} \cmidrule{4-4} \cmidrule(lr){5-6}
     & DSEC & DDD17 & MVSEC & Gen1 & 1Mpx \\
    \midrule
    \multicolumn{6}{l}{\textit{Training time (in hours). Training settings are in \cref{tab:all-config}.}}  \\
    \rowcolor{gray!20} \text{No} & 12.3 & 6.1 & 1.7 & 104.3 & 120.0 \\ 
    \text{Yes} & 23.1 & 14.8 & 4.6 & 138.9 & 155.2 \\     
    \midrule
    \multicolumn{6}{l}{\textit{Inference time on 1$\times$ RTX6000 (in milli-seconds).}}  \\
    \rowcolor{gray!20} \text{No} & 13.0 & 12.7 & 12.4 & 18.7 & 19.1 \\ 
    \text{Yes} & 14.6 & 14.0 & 13.8 & 20.4 & 20.9 \\     
    \bottomrule
  \end{tabular}
  \vspace{\tablecapmargin}
  \caption{
    Training and inference time for downstream datasets.
  }
  \label{tab:supp-downstream-runtime}
\end{table}

\subsection{Pre-training Runtime \& Model Size Analysis}

As shown in \cref{tab:supp-pretrain-runtime}, \algoNameFull only adds a marginal overhead in pre-training runtime.
Accumulating events to estimated intensity videos can be implemented efficiently on GPUs, taking less than 1 ms at each step. Detailed training and inference runtime on downstream tasks for both feed-forward and recurrent backbones are provided in \cref{tab:supp-downstream-runtime}.

We report model size of our backbone and task-specific heads in \cref{tab:supp-modulesize}.
Event-based SSL baselines such as ECDP~\cite{ECDP} and ECDDP~\cite{ECDDP} use the same backbone as ours.
However, the model size for their heads is not reported in the papers, and their open-source codebase only implements the semantic segmentation task.
Thus, we only know their segmentation head is $3\times$ larger than ours.

\section{Limitations and Future Work}  

Our approach demonstrates significant improvement across several downstream tasks. 
However, the recurrent model training requires processing of multiple stages in each iteration, which increases the training time and memory consumption.
Nevertheless, our recurrent backbone only adds lightweight recurrent modules over its feedforward counterpart, resulting in comparable inference time. 

One of the future directions is to explore alternative recurrent architectures that are better suited for event data. 
Prior work~\cite{SSMforEvents} has demonstrated that state space models~\cite{SSM} can achieve strong performance with smaller architectures. 
These characteristics offer a promising direction to reduce training time while maintaining competitive results. 

Another potential direction is the use of more diverse datasets. 
As shown in our experiments, the 1Mpx~\cite{1Mpx} dataset proves to be a better candidate than the Gen1~\cite{Gen1} dataset for pretraining. The 1Mpx dataset features a higher resolution and greater number of moving objects, which result in more diverse motion patterns. 
However, the 1Mpx dataset is limited to autonomous driving scenarios. 
We believe that a dataset containing a broader range of movement scenarios could enable the model to extract more generalizable representations from the environment.

{
    \small
    \bibliographystyle{ieeenat_fullname}
    \bibliography{main}
}

\end{document}